\documentclass[10pt,twocolumn,letterpaper]{article}
\usepackage[pagenumbers]{wacv}
\usepackage{colortbl} 
\usepackage{booktabs}
\usepackage{siunitx}
\usepackage{caption}
\usepackage{threeparttable}     
\usepackage{tabularx}
\usepackage[table,dvipsnames]{xcolor}
\usepackage{multirow}

\usepackage{pifont}
\newcommand{\cmark}{\textcolor{green!60!black}{\ding{51}}} 
\newcommand{\xmark}{\textcolor{red!70!black}{\ding{55}}}   
\usepackage[accsupp]{axessibility} 
\usepackage{tikz}
\usetikzlibrary{positioning, arrows.meta, calc, fit, backgrounds}
\usepackage{makecell}

\newlength{\maxbar}

\definecolor{iccvblue}{rgb}{0.21,0.49,0.74}
\usepackage[pagebackref,breaklinks,colorlinks,allcolors=iccvblue,plainpages=false,pdfpagelabels]{hyperref}

\def\confYear{2027}
\definecolor{lightblue}{RGB}{220,230,250}

\usepackage{fontawesome5}
\title{SignSeek: Learning Transferable Representations for Sign Dictionary Retrieval}

\author{Sobhan Asasi, Ozge Mercanoglu Sincan, Richard Bowden \\
CVSSP, University of Surrey\\
\texttt {\{s.asasi, o.mercanoglusincan, r.bowden\}@surrey.ac.uk} \\
{\small
Project Page: \url{https://cogvis-cvssp.github.io/papers/signseek/}}
}

\begin{document}
\maketitle
\addtocontents{toc}{\protect\setcounter{tocdepth}{-1}}
\vspace{-0.5em}
\begin{abstract}
Sign language dictionaries are essential resources for 
sign language learners, yet automatically retrieving a sign from a dictionary, given only
a query video, remains a challenging problem due to the natural variability between signers. Existing sign representation
learning methods are built for closed-set recognition, producing embeddings
that do not generalise to the open-set, signer-independent setting that retrieval demands.
\textbf{SignSeek} closes this gap by contrastively learning sign representations with saliency-guided articulator masking. A contrastive objective aligns same-gloss signs across signers, while our Articulator Saliency-Guided Masking (ASGM) pinpoints the single most critical articulator per sign. This drives two complementary objectives, a masked contrastive alignment (MAC) loss that sees the sign through a single articulator and a masked prediction (MAP) loss that reconstructs it in latent space from the surrounding spatio-temporal context.
Pretrained on 266K samples ($\sim$5,700 glosses) across multiple sign languages, \textbf{SignSeek} sets a new state-of-the-art performance in cross-corpus retrieval on ASL-Citizen, WLASL, and NMFs-CSL without any downstream fine-tuning.
Strikingly, it achieves zero-shot generalisation to an entirely unseen British Sign Language (BSL), surpassing methods explicitly trained on BSL, and transfers seamlessly to isolated sign recognition and subtitle alignment, outperforming prior skeleton-based methods.
\end{abstract}

\vspace{-10pt}
\section{Introduction}
\label{sec:intro}

Sign languages are the primary means of communication for millions of Deaf and
hard-of-hearing people worldwide~\cite{WFD2020, Ethnologue2023}. As living languages, they are often documented in
sign language dictionaries~\cite{Schembri2017British, schembri2013building, meinedgs_3, desai2023asl}, which are curated collections of signed video entries that
serve as essential resources for learners and educators.
Yet accessing these dictionaries remains a largely unaddressed challenge. A
learner who encounters an unfamiliar sign cannot simply \textit{look it up} the
way one would in a text based dictionary. In the task of \textbf{Sign
Dictionary Retrieval (SDR)}, given a query sign video, the goal is to retrieve the matching entry
from a large sign dictionary~\cite{desai2023asl} without any task-specific fine-tuning.
Sign dictionary retrieval supports sign language education, the documentation of linguistic variation, and other accessibility tools. Yet compared to the well-studied task of
isolated sign language recognition (ISLR)~\cite{wlasl, joze2019msasl, zhao2024masa, stgcn, jiang2021skeleton}, it has received little attention. Recognition frames the problem as closed-set classification over a
fixed vocabulary, an objective that encourages models to discriminate between
training classes but does not require them to generalise to unseen signers or
signs. SDR, by contrast, demands a metric embedding space where
similar signs cluster together regardless of signer identity,
recording conditions, or signing style, a strictly harder objective that
directly tests the transferability of representations.

Existing sign representation learning methods fail to meet these demands~\cite{signbert+, jiang2024signclip, hu2021signbert, zhao2023best, asasibeyond, asasi2026signet, asasi2025hierarchicalfeaturealignmentglossfree}.
Supervised recognition models learn decision boundaries, not transferable
distances~\cite{stgcn, jiang2021skeleton, NLA-SLR}. Self-supervised approaches largely treat sign video as generic
temporal data~\cite{hu2021signbert, signbert+, zhao2023best, wong2025signrep, sincan2025gloss}, ignoring the multi-part structure of sign language. Handshape,
arm trajectory, and non-manual features (\eg facial expressions) are not
interchangeable modalities, as each carries distinct linguistic content~\cite{SandlerLilloMartin2006, KlimaBellugi1979, afa}.
A representation that fails to capture the structural relationships between these articulators, or collapses when one is partially occluded, will fail under the variable conditions that real-world retrieval demands.
To explicitly capture this nuanced articulator structure while maintaining robustness, skeleton pose presents an ideal modality. By abstracting away background clutter and subject appearance, the pose of keypoints provides an efficient, signer-invariant representation that naturally supports cross-signer generalisation that SDR demands~\cite{jiang2021skeleton, gan2021skeleton}.

\begin{figure*}[t]
  \centering
   \includegraphics[width=\linewidth]{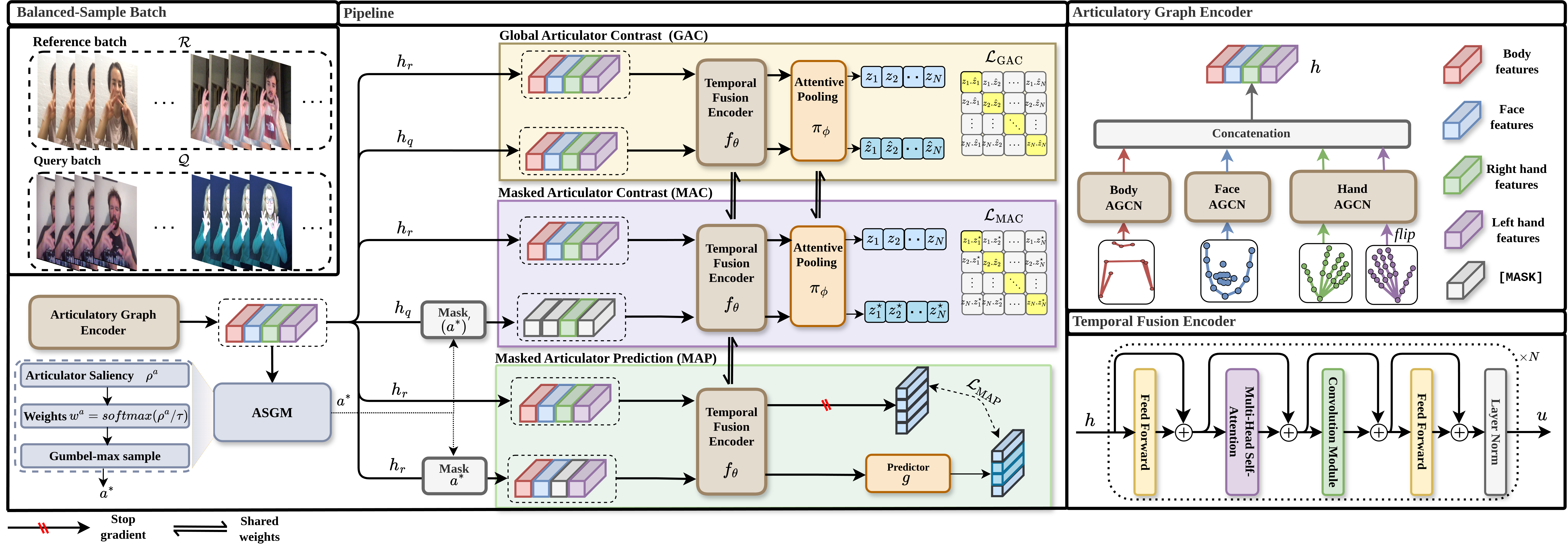}
   \captionsetup{skip=1pt, belowskip=-16pt}
   \caption{\textbf{Overview of SignSeek.} A balanced-sample batch of query and
reference clips (left) is encoded per articulator by the Articulatory Graph Encoder
(top right), fused by a Conformer $f_\theta$ (bottom right) and pooled by $\pi_\phi$.
Three objectives share this backbone. GAC contrasts intact query and reference
embeddings, MAC masks all but the dominant articulator of the query and aligns it
with clean same-gloss references, and MAP masks the dominant articulator of a
reference clip and predicts its clean representation via a predictor $g$ under a
stop-gradient. The dominant articulator $a^\star$ is selected per clip by the
Articulator Saliency-Guided Masking module (ASGM).}

   \label{fig:main}
\end{figure*}

We present \textbf{SignSeek}, a pose-based pretraining framework for sign
dictionary retrieval. 
Each articulator (face, body, hands) is encoded by its own graph network, and the resulting streams are fused by a temporal encoder, with all masking applied to the per-articulator streams before fusion. A contrastive objective aligns same-gloss samples
across signers, and our Articulator Saliency-Guided Masking (ASGM) selects the
single most critical articulator per sample and drives two complementary objectives,
a masked contrastive alignment (MAC) loss that reads the sign through this
articulator alone and a masked prediction (MAP) loss that reconstructs it in latent
space from the remaining context.
Pretrained on 266K labelled samples spanning five corpora and roughly 5,700 glosses
across Chinese, German, Turkish, and American sign languages (\cref{tab:pretrain_data}), \textbf{SignSeek} sets
a new state of the art on three retrieval benchmarks without fine-tuning
(Table~\ref{tab:retrieval}) and transfers to isolated sign recognition and subtitle
alignment (Tables~\ref{tab:wlasl_accuracy}, \ref{tab:results-how2sign}).
Figure~\ref{fig:main} overviews the framework.
Our main contributions are summarised as follows:
\textit{\textbf{(i)}} We propose \textbf{SignSeek}, a pose-based pretraining framework built on feature-level articulator masking. A saliency-guided scheme locates the most critical articulator per sample and masks it in two complementary ways, learning the sign both \emph{through} that articulator and \emph{without} it, to capture the structural dependencies between articulators.
\textit{\textbf{(ii)}} Under a strict cross-corpus protocol with no downstream fine-tuning, \textbf{SignSeek} achieves state-of-the-art retrieval on ASL-Citizen~\cite{desai2023asl}, WLASL~\cite{wlasl}, and NMFs-CSL~\cite{Hu_2021}. We also show zero-shot generalisation to an entirely unseen sign language, outperforming other approaches on British Sign Language (BSL).
\textit{\textbf{(iii)}} We show that the same representations transfer seamlessly to broader downstream tasks, surpassing prior skeleton-based methods on both isolated sign language recognition and subtitle alignment.

\section{Related Work}
\label{sec:related}

\noindent\textbf{Sign Language Representation.}
Early approaches to isolated sign language recognition (ISLR) relied on RGB
video features extracted with action-recognition backbones such as
I3D~\cite{Albanie2020bsl1k, wlasl, joze2019msasl, prajwal2022weakly, raude2024tale}, with more recent RGB methods
improving recognition through part-aware and language-assisted
modelling~\cite{NLA-SLR,shen2024stepnet,Hu_2021, ovodov2025logos}.
Pose-based methods instead operate on skeleton keypoints, providing a compact,
signer-invariant alternative to raw
pixels~\cite{hu2021signbert, zhao2023best, jiang2021skeleton, gan2021skeleton}.
Graph convolutional networks (GCNs), popularised by their success in
skeleton-based action recognition~\cite{stgcn, liu2020disentangling, shi2019skeleton, shi2019two},
are a natural fit given the graph-like organisation of the body and hands. 
Building on this, several methods pretrain GCN representations via masked joint modelling,
progressively incorporating hand, body, and facial
cues~\cite{hu2021signbert, signbert+, zhao2023best}.
However, these methods target closed-set recognition and do not yield embeddings suitable
for retrieval.
More recently, large-scale data collection has enabled transferable sign
representations that generalise across signers and datasets, typically by
mapping RGB clips to sign priors or aligning sign embeddings with gloss
text~\cite{wong2025signrep, jiang2024signclip, mslu, rust2024towards}.
In contrast, \textbf{SignSeek} operates purely on pose, needs only gloss labels, and models articulatory structure to build a metric space for open-set retrieval.

\noindent\textbf{Sign Dictionary Retrieval (SDR).}
Sign dictionary retrieval matches an isolated sign query to a gloss-level
dictionary entry using visual information alone. Despite its practical importance, 
this task has received surprisingly little direct attention. It should not be confused 
with the related task of \emph{sign language retrieval (SLRet)}. SLRet matches a 
sentence-level signing clip to a free-form text query over continuous corpora and 
requires paired sign-text data for training~\cite{chen2025c, cheng2023cico, duarte2022sign}.
Most prior work treats dictionary retrieval as a by-product of recognition,
reporting nearest-neighbour accuracy in a recognition model's embedding space rather
than optimising for retrieval directly~\cite{desai2023asl}, or listing it only as a
secondary metric within broader evaluations~\cite{wong2025signrep, mslu}.
\textbf{SignSeek} instead treats sign dictionary retrieval as a primary task, benchmarking three diverse datasets.

\begin{table}[t]
\centering
\footnotesize
\setlength{\tabcolsep}{4pt}
\renewcommand{\arraystretch}{1.15}
\resizebox{\columnwidth}{!}{%
\begin{tabular}{l l l l l}
\toprule
\textbf{Corpus} & \textbf{Language} & \textbf{Type} & \textbf{Glosses} & \textbf{Samples} \\
\midrule
MeinDGS~\cite{meinedgs_3}   & German (DGS)   & Continuous$^\dagger$ & 2{,}301 & 171{,}695 \\
SemLex~\cite{kezar2023sem}        & American (ASL) & Isolated             & 2{,}530 &  40{,}169 \\
AUTSL~\cite{Sincan_2020}          & Turkish (TİD)  & Isolated             &   226   &  28{,}142 \\
SLR500~\cite{zhao2023best}        & Chinese (CSL)  & Isolated             &   500   &  24{,}998 \\
MSASL$^\ddagger$~\cite{joze2019msasl}  & American (ASL) & Isolated             &   173   &   1{,}911 \\
\midrule
\textbf{Total} & 4 languages & Mixed & \textbf{5{,}730} & \textbf{266{,}915} \\
\bottomrule
\end{tabular}}
\captionsetup{skip=2pt, belowskip=-18pt}
\caption{\textbf{Multilingual pretraining data.} A collection combining
isolated-sign datasets with the continuous MeinDGS corpus, spanning four sign
languages with no gloss overlap. $^\dagger$MeinDGS glosses are segmented from
continuous signing. $^\ddagger$Glosses shared with WLASL (and their samples) are
removed from MSASL.}
\label{tab:pretrain_data}
\end{table}

\noindent\textbf{Sign-Subtitle Alignment (SSA).}
A key application for any sign representation is temporally aligning weakly
labelled data, such as matching subtitle text to the corresponding signing in
continuous video. Prior work typically segments the continuous stream and then
aligns the resulting segments using audio-derived timings or learned
segmentation models~\cite{moryossef2023linguistically,jiang2026segment}. More recent methods build on this by embedding
each segment and aligning it within a shared sign-text space~\cite{jiang2024signclip,jiang2026segment}.
Some train a model specifically for alignment~\cite{bull2021aligning,jang2026deep},
while pipeline methods reuse a general representation and recover the alignment via dynamic programming.
\textbf{SignSeek} follows this second paradigm, reusing its isolated-sign retrieval embeddings to align subtitles through dynamic programming without alignment-specific training.

\noindent\textbf{Contrastive and Masked Representation Learning.}
Self-supervised representation learning is largely driven by contrastive objectives and masked modelling paradigms. Supervised contrastive learning generalises instance-level contrastive objectives such as SimCLR and MoCo~\cite{khosla2020supervised,chen2020simple, he2020momentum, oord2018representation, chen2020big, chen2020improved, chen2021empirical}. It treats all samples of a class as positives, yielding more structured embedding spaces when labels are available. In the sign domain, contrastive objectives have mainly been used for cross-modal alignment~\cite{jiang2024signclip, gfslt-vlp, cheng2023cico}. 
Separately, masked modelling has proven to be a powerful pretraining signal. This is achieved either by reconstructing masked inputs, as seen in MAE and VideoMAE~\cite{he2022masked, tong2022videomae, xie2022simmim, wei2022masked, feichtenhofer2022masked, wang2023videomae}, or by predicting their representations in a learned latent space, as in the Joint Embedding Predictive Architecture (JEPA)~\cite{assran2023self, bardes2024revisiting, baevski2022data2vec}. These techniques have also been applied to skeleton sequences~\cite{hu2021signbert, signbert+, mao2023masked, zhu2023motionbert}.
\textbf{SignSeek} replaces global augmentation with articulator-level masking, coupling contrastive alignment over masked articulators with latent prediction of the dominant one.

\section{Methodology}
\label{sec:method}
We first describe our pose-based input (Sec.~\ref{sec:input}) and backbone (Sec.~\ref{sec:model}), then the Articulator Saliency-Guided Masking module that selects the most informative articulator per clip (Sec.~\ref{sec:asgm}), and finally the training objectives built on it (Sec.~\ref{sec:objectives}). \cref{fig:main} illustrates the full pipeline.

\subsection{Articulatory Representation}
\label{sec:input}
We operate on pose rather than raw video, representing each clip $X$ as a sequence 
of $T$ frames of $2$D body keypoints alongside their corresponding prediction confidence scores. 
These keypoints are partitioned into a set of distinct articulators
$\mathcal{A}=\{\mathrm{B},\mathrm{F},\mathrm{LH},\mathrm{RH}\}$, denoting the body,
face, left hand, and right hand.
For a given articulator $a\in\mathcal{A}$ with $J_a$ joints, we define its keypoint tensor
as $X^a\in\mathbb{R}^{T\times J_a\times 3}$. The final dimension of this tensor captures the 
$x$ and $y$ spatial coordinates as well as the joint estimation confidence. The complete clip is 
therefore represented as the collection $X=\{X^a\}_{a\in\mathcal{A}}$. Treating these articulators 
as distinct streams enables the model to reason about each 
component independently during the subsequent masking and fusion stages.

\subsection{Articulatory Backbone}
\label{sec:model}

\noindent\textbf{Articulatory Graph Encoder.}
Each articulator stream $X^a\in\mathbb{R}^{T\times J_a\times 3}$ (2D keypoints
with per-joint confidence) is encoded independently. A shared per-joint
projection lifts each keypoint to $C_{\text{in}}$ channels, followed by $L$
adaptive graph convolutional blocks that widen the features to $C=6C_{\text{in}}$.
Each block couples an adaptive spatial graph convolution, whose adjacency is
learned rather than fixed to the skeleton, with a temporal convolution and a
squeeze-and-excitation block that recalibrates channels~\cite{shi2019two, hu2018squeeze}.
The final block yields a joint-wise map $Z^{(L)}_t\in\mathbb{R}^{J_a\times C}$
per frame. Instead of pooling the joints, we flatten it and project back to $C$
channels with a learned matrix $W_{\text{agg}}\in\mathbb{R}^{C\times J_aC}$, $h^a_t=W_{\text{agg}}\,\mathrm{vec}\big(Z^{(L)}_t\big)\in\mathbb{R}^{C}$.
Stacking these per-frame vectors over $t=1,\dots,T$ gives the articulator stream
$h^a\in\mathbb{R}^{T\times C}$, so the aggregation learns a per-joint weighting
rather than treating all joints equally. The body and face are each processed by
their own dedicated encoder, while the two hands share a single encoder
$E_{\text{H}}$ since they are articulatorily equivalent, with the right hand
mirrored by an $x$-axis flip $\Phi_x$ so both are seen in a common canonical
frame, then we concatenate each encoder's output (\cref{fig:main},~top~right) :
\begin{equation}
\begin{aligned}
h^{\mathrm{B}}&=E_{\mathrm{B}}(X^{\mathrm{B}}),\ \ \ \ \quad
h^{\mathrm{F}}=E_{\mathrm{F}}(X^{\mathrm{F}}),\\
h^{\mathrm{RH}}&=E_{\mathrm{H}}(X^{\mathrm{RH}}),\quad
h^{\mathrm{LH}}=E_{\mathrm{H}}\big(\Phi_x(X^{\mathrm{LH}})\big),\\
h&=\big[\,h^{\mathrm{B}}\,\|\,h^{\mathrm{F}}\,\|\,h^{\mathrm{RH}}\,\|\,h^{\mathrm{LH}}\,\big]\in\mathbb{R}^{T\times NC}, \  N=|\mathcal{A}|.
\end{aligned}
\end{equation}

\noindent\textbf{Temporal Fusion Encoder.}
The concatenated stream is then
linearly projected to the model dimension $d_m$ and passed to a Conformer encoder, whose architecture is shown in the bottom-right of \cref{fig:main},
that captures long-range temporal structure by interleaving multi-head
self-attention with convolution, producing a contextualised sequence:
\begin{equation}
\mathbf{s}=f_\theta(h)\in\mathbb{R}^{T\times d_m}.
\end{equation}
An attentive pooling head $\pi_\phi$ then maps this sequence to a single clip
embedding, reusing the Conformer's self-attention to weight frames before a
two-layer projection:
\begin{equation}
z=\pi_\phi(\mathbf{s})\in\mathbb{R}^{d}.
\end{equation}

We L2-normalise $z$ and use it both for the contrastive objectives of
Sec.~\ref{sec:objectives} and, at inference, as the retrieval embedding.

\begin{table*}[t]
  \centering
  \footnotesize
  \renewcommand{\arraystretch}{1.0}
  \setlength{\tabcolsep}{5.5pt}
  \begin{tabular}{l| c| c ccc c ccc c ccc}
    \toprule
    \multirow{2}{*}{\textbf{Features}} & \multirow{2}{*}{\textbf{Dataset (\textit{Scale})}} & & \multicolumn{3}{c}{\textbf{ASL-Citizen}}
    & & \multicolumn{3}{c}{\textbf{WLASL2000}}
    & & \multicolumn{3}{c}{\textbf{NMFs-CSL}} \\
        \cmidrule(lr){4-6} \cmidrule(lr){8-10} \cmidrule(lr){12-14}
    & &
      & {DCG\,$\uparrow$} & {R@1\,$\uparrow$} & {R@5\,$\uparrow$} &
      & {DCG\,$\uparrow$} & {R@1\,$\uparrow$} & {R@5\,$\uparrow$} &
      & {DCG\,$\uparrow$} & {R@1\,$\uparrow$} & {R@5\,$\uparrow$} \\

    \midrule
    \rowcolor{gray!17}
    \multicolumn{14}{@{}l}{\textbf{Image / video pre-trained features}}\\
    \midrule
    HieraMAE~\cite{ryali2023hiera, wong2025signrep} & Kinetics-400 (700 \textit{h})      & & 11.64 &  0.25 &  0.84 & & 13.21 & 2.08 &  3.40 & & 23.29 &  3.96 & 12.18 \\
    HieraMAE~\cite{ryali2023hiera, wong2025signrep}    & YT-SL-25 (3,000 \textit{h})         & & 12.12 &  0.39 &  1.34 & & 14.06 & 2.57 &  4.41 & & 28.03 &  7.57 & 18.38 \\
    I3D~\cite{Albanie2020bsl1k} & BSL-1K (273 \textit{K}) && 13.32 & 0.63 & 2.06 &  & 17.06 & 3.41 & 6.81 &  & 25.82 & 6.34 & 16.03 \\
    Video-Swin~\cite{prajwal2022weakly} & BOBSL (5,500 \textit{K}) && 27.21 & 7.55 & 19.38 &  & 36.84 & 13.97 & 33.39 &  & 65.31 & 43.97 & 71.62 \\
    \midrule
    \rowcolor{gray!17}
    \multicolumn{14}{@{}l}{\textbf{Pose / keypoint features}}\\
    \midrule
    All Joint Angles~\cite{wong2025signrep}  & \textbf{{--}}               & & 13.82 &  0.57 &  2.17 & & 17.92 & 2.54 &  6.50 & & 32.51 &  7.93 & 25.50 \\
    All Keypoints~\cite{wong2025signrep}     & \textbf{{--}}               & & 14.29 &  0.98 &  2.74 & & 19.37 & 3.16 &  8.23 & & 36.64 & 12.26 & 30.99 \\
    Hand Keypoints~\cite{wong2025signrep}    & \textbf{{--}}               & & 25.91 &  7.96 & 19.58 & & 28.11 & 7.57 & 20.92 & & 41.72 & 15.91 & 42.62 \\
    Hand Joint Angle~\cite{wong2025signrep}  & \textbf{{--}}               & & 26.93 &  8.81 & 21.41 & & 30.61 & 9.42 & 24.36 & & 44.17 & 18.13 & 46.34 \\
    \midrule
    \rowcolor{gray!17}
    \multicolumn{14}{@{}l}{\textbf{Learned sign representations}}\\
    \midrule
    SignCLIP~\cite{jiang2024signclip}         & Spreadthesign (456 \textit{K})      & & 21.80 &  4.00 & 11.42 & & 22.10 & 5.25 & 11.78 & & 51.20 & 26.75 & 53.03 \\
    MASA~\cite{zhao2024masa} & 4$\times$ ISLR sets (146 \textit{K}) & 
    & 49.31 & 27.96 & 49.78
    & & 36.98\rlap{$^\dag$} & 14.38\rlap{$^\dag$} & 32.63\rlap{$^\dag$}
    & & 74.94\rlap{$^\dag$} & 54.69\rlap{$^\dag$} & 83.27\rlap{$^\dag$} \\
    SignRep~\cite{wong2025signrep}          & YT-SL-25 (3,000 \textit{h})             & & \underline{71.21} & \underline{49.95} & \underline{80.09}
                     & & \underline{57.93} & \underline{29.92} & \underline{67.41}
                     & & \underline{83.05} & \underline{63.04} & \underline{95.63} \\
    \midrule
    \rowcolor{lightblue}
    \textbf{SignSeek (Ours)} & 5$\times$ corpora~(266 \textit{K})     &
      & \bfseries 77.21 & \bfseries 57.66 & \bfseries 86.67
      & & \bfseries 61.24 & \bfseries 34.75 & \bfseries 70.25
      & & \bfseries 85.17 & \bfseries 65.20 & \bfseries 96.79 \\
    \bottomrule
  \end{tabular}
  \captionsetup{skip=1pt, belowskip=-15pt}
\caption{\textbf{Cross-corpus sign dictionary retrieval.} No method is fine-tuned on
the evaluation datasets. The evaluation datasets are unseen during pre-training
except where marked $\dagger$, which indicates the dataset was part of that method's
pre-training set (MASA includes WLASL and NMFs-CSL). \textbf{SignSeek} is pretrained
on five sign corpora across four languages (Table~\ref{tab:pretrain_data}), none
overlapping the evaluation sets. \textbf{Best} in bold, \underline{second best}
underlined. Scale is in hours (\textit{h}) for continuous-signing corpora and thousands of clips
(\textit{K}) for isolated-sign sets.
}

  \label{tab:retrieval}
\end{table*}

\subsection{Articulator Saliency-Guided Masking (ASGM)}
\label{sec:asgm}
Our masking-based objectives depend on knowing which articulator carries each sign, which we derive from the encoder itself rather than a fixed rule.
We use the norm of each articulator's temporally pooled embedding as a simple proxy
for its salience, since the more the model relies on an articulator, the larger the
norm of its embedding.
The per-articulator encoders are separately parameterised and have different joint
counts, however, their raw feature scales are not directly comparable. We therefore
standardise each stream by its running per-channel mean $\mu^a$ and standard
deviation $\sigma^a$ before scoring, and take the norm of the standardised,
temporally pooled representation,
\begin{equation}
\bar h^a=\frac{1}{|\mathcal{T}|}\sum_{t\in\mathcal{T}}h^a_t,\qquad
\rho^a=\left\|\,\frac{\bar h^a-\mu^a}{\sigma^a}\,\right\|_2,\qquad a\in\mathcal{A},
\end{equation}
where $\mathcal{T}$ denotes the valid (non-padded) frames, the standardisation is
per channel, and $(\mu^a,\sigma^a)$ are running statistics updated during training.
This makes $\rho^a$ measure activation relative to each stream's own baseline, so the
scores are commensurable across articulators, and it needs no auxiliary supervision
and no learned parameters. A temperature-scaled softmax then converts the scores into
a distribution over articulators,
\begin{equation}
w^a=\frac{\exp(\rho^a/\tau)}{\sum_{a'\in\mathcal{A}}\exp(\rho^{a'}/\tau)},
\end{equation}
in which the temperature $\tau$ modulates how sharply the mass concentrates on the
dominant articulator, interpolating between near-uniform exploration and a
near-deterministic focus. Committing to a hard $\arg\max$ would collapse this
distribution and starve the objectives of variety, so we instead sample the
dominant articulator $a^\star$ with the Gumbel-max reparameterisation:
\begin{equation}
a^\star=\arg\max_{a\in\mathcal{A}}\big(\log w^a+\gamma^a\big), \quad \gamma^a=-\log(-\log u^a),
\end{equation}
with $u^a\sim\mathcal{U}(0,1)$ drawn independently per articulator, which yields
exactly one articulator per clip while preserving the saliency ordering in
expectation. 
This single choice $a^\star$ then anchors two complementary objectives
(Sec.~\ref{sec:objectives}), one that learns the sign \emph{through} $a^\star$ alone
and one that learns it \emph{without} $a^\star$, turning masking into a structured
learning signal rather than mere augmentation.

\subsection{Training Objectives}
\label{sec:objectives}

\noindent\textbf{Balanced Batch Sampling.}
We sample each batch as $k$ glosses with $n$ clips each, split into a reference batch
and a query batch of different clips of the same glosses, so every query has
same-gloss references as positives. A reference clip with streams $h_r$ is encoded
into an intact embedding $z=\pi_\phi\big(f_\theta(h_r)\big)$, and a query clip $h_q$
into an embedding set by the objective's masking, defined below. When an objective
masks an articulator, it replaces the corresponding stream by a learned
per-articulator token $m^{a}$ before fusion.

\noindent\textbf{Global Articulator Contrast (GAC).}
GAC is our base objective, a supervised contrastive loss~\cite{khosla2020supervised}
over intact clip embeddings with no masking. 
With no masking, the reference and query clips give intact embeddings $z_i$ and
$\hat z_i$. Let $I=\{z_i\}_i\cup\{\hat z_i\}_i$ collect them over the batch, and let
$P(v)$ denote those sharing the gloss of $v$.
The loss pulls same-gloss embeddings together and pushes
the rest apart,
\begin{equation}
\mathcal{L}_{\text{GAC}}=\sum_{v\in I}\frac{-1}{|P(v)|}
\sum_{p\in P(v)}\log
\frac{\exp\!\big(v^{\top}p/\tau_c\big)}
{\sum_{j\in I\setminus\{v\}}\exp\!\big(v^{\top}j/\tau_c\big)},
\label{eq:gac}
\end{equation}
with temperature $\tau_c$ and L2-normalised embeddings. Because positives are
defined by gloss rather than by instance, same-sign embeddings from different clips
and signers are drawn together, shaping the signer-invariant space that retrieval
demands.

\noindent\textbf{Masked Articulator Contrast (MAC).}
MAC introduces saliency-guided masking into the contrastive objective. Guided by
the dominant articulator $a^{\star}$ from ASGM, it keeps only that stream and
replaces every other by the mask token $m^{a}$,
\begin{equation}
z^{\star}=\pi_\phi\big(f_\theta(h_q^{\star})\big),\qquad
h_q^{\star,a}=\begin{cases}h_q^{a} & a=a^{\star}\\[2pt] m^{a} & a\neq a^{\star}\end{cases}
\label{eq:mac_mask}
\end{equation}
so that the sign must be read through its most salient articulator alone. Let
$M=\{z_i\}_i\cup\{z^\star_i\}_i$ collect the reference embeddings $z_i$ and the
dominant-only query embeddings $z^\star_i$, and let $P(u)$ denote those sharing the
gloss of $u$.
MAC pulls same-gloss
embeddings together and pushes the rest apart,
\begin{equation}
\mathcal{L}_{\text{MAC}}=\sum_{u\in M}\frac{-1}{|P(u)|}
\sum_{p\in P(u)}\log
\frac{\exp\!\big(u^{\top}p/\tau_c\big)}
{\sum_{j\in M\setminus\{u\}}\exp\!\big(u^{\top}j/\tau_c\big)},
\label{eq:mac}
\end{equation}
with the same temperature $\tau_c$ and L2-normalised embeddings as in
Eq.~\eqref{eq:gac}. Because the dominant-only query must still align with clean same-gloss references, the most salient articulator alone is forced to be gloss-discriminative.

\noindent\textbf{Masked Articulator Prediction (MAP).}
Unlike the contrastive GAC and MAC, MAP operates within a single reference clip and
adds no negatives. Both of its branches take a reference clip $h_r$, one masks the
dominant stream $a^{\star}$ to form a masked context, the other encodes the clip
intact as a prediction target. The masked context sequence is
\begin{equation}
\mathbf{s}^{\star}=f_\theta(h_r^{\dagger})\in\mathbb{R}^{T\times d_m},\qquad
h_r^{\dagger,a}=\begin{cases}m^{a} & a=a^{\star}\\[2pt] h_r^{a} & a\neq a^{\star}\end{cases}
\label{eq:map_mask}
\end{equation}
A predictor $g:\mathbb{R}^{d_m}\!\to\!\mathbb{R}^{d_m}$, a lightweight transformer
decoder that attends over the whole masked sequence, predicts the clean Conformer
sequence $\mathbf{s}=f_\theta(h_r)$ of the same clip, held fixed by a stop-gradient
$\mathrm{sg}[\cdot]$ so gradients flow only through the predictor and the online
encoder. This latent target avoids the collapse that raw-coordinate reconstruction
can induce. MAP maximises the frame-wise alignment between prediction and target,
\begin{equation}
\mathcal{L}_{\text{MAP}}=\mathbb{E}_{t\in\mathcal{T}}
\left[\,1-
\frac{\big\langle g(\mathbf{s}^{\star})_t,\;\mathrm{sg}\big[\mathbf{s}_t\big]\big\rangle}
{\big\|g(\mathbf{s}^{\star})_t\big\|_2\,\big\|\mathrm{sg}\big[\mathbf{s}_t\big]\big\|_2}
\,\right],
\label{eq:map}
\end{equation}
averaged over the valid frames $\mathcal{T}$. By recovering from the surviving
context the representation the dominant articulator would have contributed, MAP
captures the structural dependencies between articulators that underlie sign meaning.

\noindent\textbf{Joint Objective.}
We optimise the three terms jointly,
\begin{equation}
\mathcal{L}=
\lambda_{\text{GAC}}\,\mathcal{L}_{\text{GAC}}
+\lambda_{\text{MAC}}\,\mathcal{L}_{\text{MAC}}
+\lambda_{\text{MAP}}\,\mathcal{L}_{\text{MAP}},
\label{eq:total}
\end{equation}
where $\lambda_{\text{GAC}}$, $\lambda_{\text{MAC}}$, and $\lambda_{\text{MAP}}$
balance the three objectives.

\noindent\textbf{Inference.}
All masking, the ASGM module, and the predictor $g$ are training-time only. At
inference we discard them and pass a clip through the backbone unchanged, from the
articulatory graph encoders through the temporal fusion Conformer and the attentive
pooling head, giving a single embedding $z=\pi_\phi\big(f_\theta(h)\big)$ from the
intact streams $h$. Retrieval then ranks dictionary entries by cosine similarity
between the L2-normalised query and reference embeddings, with no downstream
fine-tuning.

\section{Experiments}
\label{sec:experiments}

\subsection{Datasets and Evaluation Protocol}
\label{sec:protocol}

\noindent\textbf{Training data.}
We pre-train on five sign corpora across four languages, combining isolated-sign datasets with gloss-segmented clips from the
continuous MeinDGS corpus, totalling 266K labelled clips and roughly 5{,}700 glosses
(Table~\ref{tab:pretrain_data}). The encoder is trained once and then frozen for all
evaluations.

\noindent\textbf{Evaluation benchmarks.}
We evaluate sign dictionary retrieval on three datasets that are disjoint from the training corpus, ASL-Citizen~\cite{desai2023asl}, WLASL~\cite{wlasl}, and
NMFs-CSL~\cite{Hu_2021}. Because none of them are seen during
pre-training and no parameters are fine-tuned on them, all results are zero-shot.
Beyond these benchmarks, we further study a dictionary-lookup scenario
for British Sign Language, a language entirely unseen during pre-training. As shown
in \cref{tab:retrieval-bsl}, we query a BSL evaluation set of $726$
query--gloss pairs, generated by a native BSL signer, against the publicly available BSL SignBank dictionary~\cite{signbank}.
This mirrors how a user would search an unfamiliar sign against a reference lexicon
and tests cross-lingual transfer of the model to a new vocabulary. Results show our pose encoder
retrieves the correct gloss significantly more often than prior work despite never
having been trained or fine-tuned on BSL.
Finally, we apply the same frozen encoder to sign-subtitle alignment on
How2Sign~\cite{how2sign} and BOBSL~\cite{bobsl}, again without any fine-tuning.

\noindent\textbf{Retrieval protocol.}
We treat the test split as queries and the training split as the dictionary gallery. Every
clip is mapped to an $\ell_2$-normalised embedding, and each gallery gloss is
scored by the maximum cosine similarity between the query and that gloss'
instances. Glosses are then ranked by this score, and we report metrics on the
rank $r$ of the correct gloss, averaged over queries. From this ranking we report Recall@$k$ (R@$k$, the fraction of queries whose correct
gloss appears in the top $k$), mean average precision (mAP), and discounted
cumulative gain (DCG)~\cite{jarvelin2002cumulated, desai2023asl}.

\noindent\textbf{Downstream metrics.}
For isolated sign language recognition we report Top-$1$ and Top-$5$ accuracy,
both per instance and per class. For sign-subtitle alignment we report frame-level
F1 at an intersection-over-union threshold of $0.5$ (F1@$0.5$).

\begin{table}[t]
\centering
\footnotesize
\renewcommand{\arraystretch}{1.2}
\setlength{\tabcolsep}{3pt}
\begin{tabular}{l| c c c| ccc}
\toprule
\textbf{Model} & \textbf{Input} & \textbf{\#P (\textit{M})} & \shortstack{\textbf{BSL}\\\textbf{Seen?}} & \textbf{R@1}$\uparrow$ & \textbf{R@5}$\uparrow$ & \textbf{mAP}$\uparrow$ \\
\midrule
I3D~\cite{Albanie2020bsl1k}         & RGB  & 13.4  & \cmark & 0.55  & 2.22  & 1.67 \\
SignCLIP~\cite{jiang2024signclip}   & Pose & 109.4 & \xmark & 1.02  & 3.05  & 2.85 \\
SignRep~\cite{wong2025signrep}      & RGB  & 51.5  & \xmark & 4.30  & 12.62 & 8.72 \\
MASA~\cite{zhao2024masa} 
 & Pose  & 68.2  & \xmark & 6.80  & 15.40 & 11.57 \\
Video-Swin~\cite{prajwal2022weakly}         & RGB & 49.5  & \cmark & \underline{9.15}  & \underline{16.78} & \underline{13.37} \\
\midrule
\rowcolor{lightblue}
 &  &  &  & \textbf{12.76} & \textbf{29.54} & \textbf{20.84} \\
\rowcolor{lightblue}
\multirow{-2}{*}{\textbf{SignSeek (Ours)}} & \multirow{-2}{*}{Pose} & \multirow{-2}{*}{15.7} & \multirow{-2}{*}{\xmark}
 & \textcolor{ForestGreen}{\textbf{(+3.61)}} & \textcolor{ForestGreen}{\textbf{(+12.76)}} & \textcolor{ForestGreen}{\textbf{(+7.47)}} \\
\bottomrule
\end{tabular}
\captionsetup{skip=2pt, belowskip=-7pt}
\caption{\textbf{Zero-shot BSL dictionary retrieval.} We query a BSL
evaluation set ($726$ query--gloss pairs) against the BSL SignBank
dictionary and report. \textbf{Input}: RGB video or body
pose. \textbf{\#P}: parameters (M) of the frozen feature extractor.
\textbf{BSL Seen?}: whether the backbone observed British Sign Language during
pre-training. 
No method is fine-tuned on SignBank. }
\label{tab:retrieval-bsl}
\end{table}

\begin{figure}[t]
    \centering
    \footnotesize
    \includegraphics[width=0.8\linewidth]{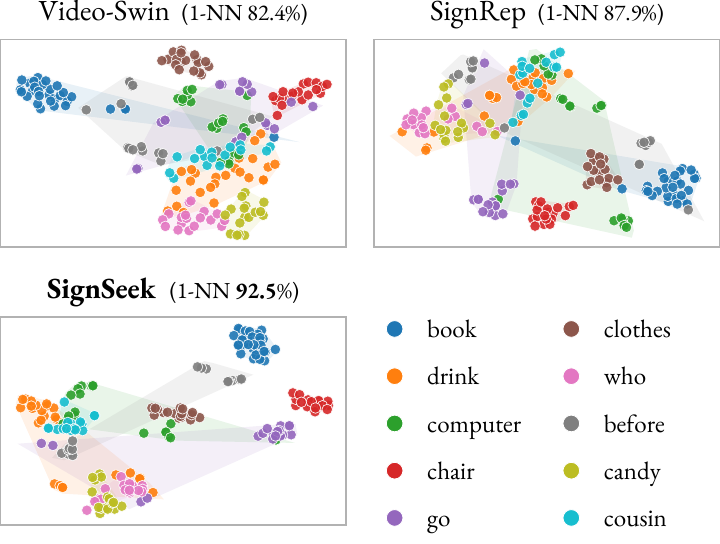}
    \captionsetup{skip=2pt, belowskip=-15pt}
\caption{\textbf{t-SNE of WLASL embeddings}. Titles report 1-NN accuracy (\% of samples whose nearest neighbour shares the gloss).}

    \label{fig:tsne}
\end{figure}
\subsection{Implementation Details}
\label{sec:impl}

We train for $30,000$ steps with AdamW, a learning rate of
$3\times10^{-4}$, weight decay $0.01$, and $3,000$ warmup steps. Batches are drawn
by balanced sampling of $k{=}40$ glosses with $n{=}2$. 
The contrastive temperature is $\tau_c{=}0.07$, the ASGM
selection temperature is $\tau{=}0.7$, and the training objectives are weighted
equally.  Full
hyperparameters are provided in the supplementary.

\begin{figure}
    \centering
    \includegraphics[width=1\linewidth]{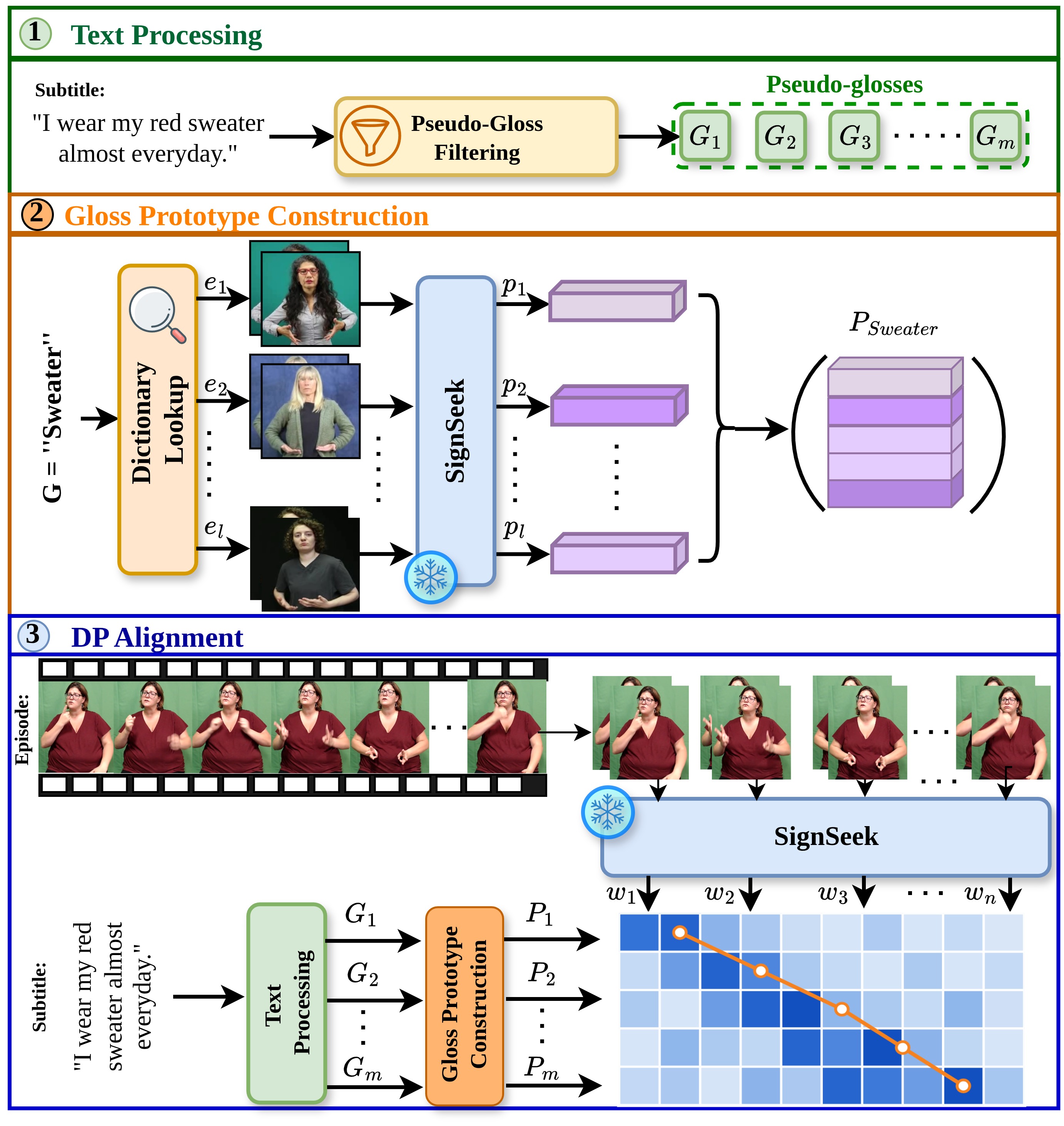}
  \captionsetup{skip=2pt, belowskip=-15pt}
    \caption{\textbf{SSA with SignSeek.} A subtitle is filtered
  into pseudo-glosses (1), each looked up in a language-matched isolated-sign
  dictionary (ASL or BSL) and encoded by the frozen SignSeek encoder into a
  prototype bank $P_G$ (2). The same encoder embeds the episode windows, and a
  monotonic path through the gloss-to-window similarity matrix is decoded by
  dynamic programming(3).}
    \label{fig:ssa-pipeline}
\end{figure}
\subsection{Results}
\label{sec:results}
\noindent\textbf{Sign Dictionary Retrieval.}
We benchmark \textbf{SignSeek} against a broad range of prior representations as
frozen feature extractors for dictionary retrieval, where no method observes the
evaluation data during pre-training and none is fine-tuned, except MASA which sees
WLASL2000 and NMFs-CSL in training (marked $\dag$ in \cref{tab:retrieval}). On ASL-Citizen, WLASL2000, and NMFs-CSL
(\cref{tab:retrieval}), image and video features transfer poorly and
hand-focused pose descriptors remain well behind the learned representations.
\textbf{SignSeek} improves over the strongest baseline, SignRep, on every metric and
dataset. It raises DCG from $71.21$ to $77.21$ on ASL-Citizen, from $57.93$ to
$61.24$ on WLASL2000, and from $83.05$ to $85.17$ on NMFs-CSL. R@1 rises
correspondingly, from $49.95$ to $57.66$, from $29.92$ to $34.75$, and from $63.04$
to $65.20$.
\textbf{SignSeek} also generalises to an unseen language. We query our native-signer
BSL evaluation set ($726$ clips) against the BSL SignBank dictionary
(\cref{tab:retrieval-bsl}), a language absent from pre-training. Although the
strongest baselines I3D and Video-Swin were trained on BSL through BSL-1K or BOBSL,
\textbf{SignSeek} reaches $12.76$ R@1 and $20.84$ mAP from body pose alone with a
frozen extractor of only $15.7$M parameters, surpassing the in-domain Video-Swin by
$3.61$ R@1 and $7.47$ mAP. This indicates the gains stem from a more transferable
representation rather than language-specific exposure or model scale.
\cref{fig:tsne} visualises the learned space with a t-SNE projection of the ten most frequent WLASL glosses. Compared with Video-Swin and SignRep,
\textbf{SignSeek} forms tighter and better-separated per-gloss clusters, reflected
in a higher 1-NN accuracy. We provide qualitative results in the supplementary.
\begin{table}[t]
    \centering
    \footnotesize
    \begin{threeparttable}
    \begin{tabularx}{\columnwidth}{X c c c c }
      \toprule
      & \multicolumn{2}{c}{\textbf{How2Sign}} & \multicolumn{2}{c}{\textbf{BOBSL}} \\
      \cmidrule(lr){2-3} \cmidrule(lr){4-5}
      \textbf{Method} & Val & Test & Val & Test \\
      \midrule
      \rowcolor{gray!17} \multicolumn{5}{l}{\textbf{Static Baselines}} \\
      \midrule
      \noalign{\vskip 1pt}
      Original (audio-based)              & 30.63 & 33.06 & 29.09 & 14.11 \\
      Original\textsuperscript{+}(fixed offsets) & 31.91 & 36.21  & 49.63 & 44.61\\
      \midrule
      \rowcolor{gray!17} \multicolumn{5}{l}{\textbf{Segment and Align}} \\
      \midrule
      \noalign{\vskip 1pt}
      Segmentation~\cite{moryossef2023linguistically} & 33.38 & 36.17 & 66.24 & 49.58 \\
      Segmentation-BSLCP~\cite{jiang2026segment} & - & - &  63.61 & 47.75 \\
      \midrule
      \rowcolor{gray!17} \multicolumn{5}{l}{\textbf{Segment, Embed, and Align}} \\
      \midrule
      \noalign{\vskip 1pt}
      SignCLIP~\cite{jiang2024signclip}  & 35.51 & 37.51 & 66.70 & 50.68\\
      \midrule
      \rowcolor{gray!17} \multicolumn{5}{l}{\textbf{Dictionary-Guided Alignment}} \\
      \midrule
      \rowcolor{lightblue}
      \textbf{SignSeek}  & \textbf{36.51} & \textbf{40.20} & \textbf{66.94} & \textbf{50.72} \\
      \bottomrule
    \end{tabularx}
    \captionsetup{skip=2pt, belowskip=-15pt}
    \captionof{table}{\textbf{Subtitle alignment.} Sentence-level
      temporal alignment accuracy (F1@0.5$\uparrow$) on the validation and test
      splits of How2Sign and BOBSL. Methods are grouped by alignment strategy, from audio-based static
      baselines to our dictionary-guided approach.}
    \label{tab:results-how2sign}
    \end{threeparttable}
\end{table}

\subsection{Sign-Subtitle Alignment (SSA)}
\label{sec:results-alignment}

\cref{tab:results-how2sign} evaluates our frozen embeddings on
sentence-level subtitle alignment. As shown in \cref{fig:ssa-pipeline}, for each pseudo-gloss $G_i$ produced by pseudo-gloss filtering from the input subtitle, \textbf{SignSeek} retrieves $l_i$ isolated-sign exemplars from a language-matched dictionary (ASL for How2Sign, BSL for BOBSL) and encodes them with the frozen encoder into a prototype set $P_i$. The same
encoder embeds the video windows $w_1,\dots,w_n$, and each entry of the similarity matrix is the maximum cosine similarity between a window and the prototypes of a gloss, $S(i,j)=\max_{p\in P_i}\langle p, w_j\rangle$. A
monotonic gloss-to-window correspondence is then decoded with dynamic
programming (\cref{fig:ssa-pipeline}). The encoder is frozen in this task, so the method aligns without any
task-specific training.
All methods in \cref{tab:results-how2sign} reuse a general representation without alignment-specific training,
recovering the alignment by dynamic programming, so they differ only in the
representation they reuse. \textbf{SignSeek} achieves the best F1@$0.5$ on How2Sign and remains
competitive with SignCLIP on BOBSL. It improves over SignCLIP by 1.0 on How2Sign
validation and 2.7 on test, and matches it on BOBSL, 66.94 to 66.70 and 50.72 to
50.68. Unlike the representations reused by the baselines, which require
continuous-signing boundary annotations or paired sign-text, \textbf{SignSeek}'s encoder is
trained from pose with only gloss labels. See the supplementary for qualitative results and further details.

\begin{table}[t]
\centering
\footnotesize
\setlength{\tabcolsep}{4pt}
\renewcommand{\arraystretch}{1}
\begin{tabularx}{\columnwidth}{X c c c c}
\toprule
 &
\multicolumn{2}{c}{\textbf{Instance Acc.}} &
\multicolumn{2}{c}{\textbf{Class Acc.}} \\
\cmidrule(lr){2-3}
\cmidrule(lr){4-5}
\textbf{Method} & Top-1 & Top-5 & Top-1 & Top-5 \\
\midrule

\rowcolor{gray!17}
\rowcolor{gray!17}
\multicolumn{5}{l}{\textbf{Skeleton-based}} \\
\midrule
ST-GCN~\cite{stgcn}     & 34.40 & 66.57 & 32.53 & 65.45 \\
SignBERT~\cite{hu2021signbert}   & 39.40 & 73.35 & 36.74 & 72.38 \\
BEST~\cite{zhao2023best}       & 46.25 & 79.33 & 43.52 & 77.65 \\
SignBERT+~\cite{signbert+}  & 48.85 & 82.48 & 46.37 & 81.33 \\
MASA~\cite{zhao2024masa} & 49.06 & 82.90 & 46.91 & 81.80 \\
\midrule

\rowcolor{lightblue}
\textbf{SignSeek}
& \textbf{56.65}
& \textbf{87.67}
& \textbf{53.64}
& \textbf{87.53} \\

\bottomrule
\end{tabularx}
\captionsetup{skip=2pt, belowskip=-10pt}
\caption{\textbf{ISLR results on WLASL.}}
\label{tab:wlasl_accuracy}
\end{table}

\begin{table}[t]
\centering
\footnotesize
\begin{tabularx}{\columnwidth}{X c c c c c c}
\toprule
& \multicolumn{2}{c}{\textbf{NMFs-CSL}}
& \multicolumn{2}{c}{\textbf{ASL-Citizen}} \\

\cmidrule(lr){2-3}
\cmidrule(lr){4-5}

\textbf{Method}
& Top-1 & Top-5
& Top-1 & Top-5 \\

\midrule
\rowcolor{gray!17}
\multicolumn{5}{l}{\textbf{Skeleton-based}} \\
\midrule

ST-GCN~\cite{stgcn}
& 59.9 & 86.8
& 59.52 & 82.68 \\

SignBERT~\cite{hu2021signbert}
& 67.0 & 95.3
& \textbf{--} & \textbf{--} \\

BEST~\cite{zhao2023best} 
& 68.5 & 94.4
& \textbf{--} & \textbf{--} \\
MASA~\cite{zhao2024masa} & 71.7 & 97.0 &  \textbf{--} & \textbf{--} \\

SignCLIP~\cite{jiang2024signclip}
& \textbf{--} & \textbf{--}
& 60.0 & 84.0 \\

\midrule

\rowcolor{lightblue}
\textbf{SignSeek}
& \textbf{73.8} & \textbf{98.8}
& \textbf{74.7} & \textbf{94.0} \\

\bottomrule

\end{tabularx}
\captionsetup{skip=2pt, belowskip=-15pt}
\captionof{table}{\textbf{ISLR results on NMFs-CSL and ASL-Citizen.}}
\label{tab:islr_nmf_asl}
\end{table}

\subsection{Isolated Sign Language Recognition (ISLR)}
\label{sec:results-islr}

To test whether the pretrained encoder transfers to recognition, we add a
classification layer on top and fine-tune the whole model. Under this protocol,
\textbf{SignSeek} is the strongest skeleton-based method on all three datasets
(\cref{tab:wlasl_accuracy}, \cref{tab:islr_nmf_asl}). On WLASL it reaches
$56.65\%$ Top-$1$ instance accuracy, $7.6$ points above the best prior skeleton
model, MASA. On NMFs-CSL it improves Top-$1$ by $2.1$ points over MASA, and on
ASL-Citizen it exceeds the nearest skeleton and text-aligned baselines by
$14.7$ points. These gains show that our pretrained representation transfers
competitively to recognition. 

\subsection{Ablation Studies}
\label{sec:ablations}

\noindent\textbf{Contribution of each objective.}
\cref{tab:ablation_retrieval} studies the three objectives, starting from the GAC
baseline. Adding MAC alone raises R@$1$ from $27.25$ to $31.27$ on WLASL and from
$48.40$ to $54.15$ on ASL-Citizen, while MAP alone gives a slightly larger gain
($32.33$ and $55.93$). Combining both on top of GAC is best, reaching $34.75$ on
WLASL and $57.66$ on ASL-Citizen, above either branch alone, so understanding a sign
both \emph{through} and \emph{without} its dominant articulator provides
complementary signals. Removing GAC while keeping both masked objectives drops
R@$1$ to $32.45$ and $55.12$, below the full model,
showing that the global contrastive anchor is needed for the masked objectives to
reach their full effect.

\newcolumntype{Y}{>{\centering\arraybackslash}X}
\begin{table}[t]
    \centering
    \footnotesize
    \setlength{\tabcolsep}{4pt}
    \begin{threeparttable}
    \begin{tabularx}{\columnwidth}{ccc YYYY}
      \toprule
      & & & \multicolumn{2}{c}{\textbf{WLASL}} & \multicolumn{2}{c}{\textbf{ASL-Citizen}} \\
      \cmidrule(lr){4-5}\cmidrule(lr){6-7}
      \textbf{GAC} & \textbf{MAC} & \textbf{MAP} & R@1$\uparrow$ & DCG$\uparrow$ & R@1$\uparrow$ & DCG$\uparrow$ \\
      \midrule
      \cmark & \xmark & \xmark & 27.25 & 54.02 & 48.40 & 70.02 \\
      \cmark & \cmark & \xmark & 31.27 & 58.55 & 54.15 & 75.02 \\
      \cmark & \xmark & \cmark & 32.33 & 60.16 & 55.93 & 75.77 \\
      \xmark & \cmark & \cmark & 32.45 & 60.20    & 55.12    & 75.05    \\
      \midrule
      \rowcolor{lightblue}
      \cmark & \cmark & \cmark & \textbf{34.75} & \textbf{61.24} & \textbf{57.66} & \textbf{77.21} \\
      \bottomrule
    \end{tabularx}
    \captionsetup{skip=2pt, belowskip=-11pt}
\captionof{table}{\textbf{Objective ablation.} Each row toggles GAC, MAC, and MAP.
The full model with all three (highlighted) is \textbf{SignSeek}.}

    \label{tab:ablation_retrieval}
    \end{threeparttable}
\end{table}

\begin{table}[t]
    \centering
    \footnotesize
    \setlength{\tabcolsep}{4pt}
    \begin{threeparttable}
    \begin{tabularx}{\columnwidth}{X cc cc}
      \toprule
      & \multicolumn{2}{c}{\textbf{WLASL}} & \multicolumn{2}{c}{\textbf{ASL-Citizen}} \\
      \cmidrule(lr){2-3} \cmidrule(lr){4-5}
      \textbf{Setting} & R@1$\uparrow$ & DCG$\uparrow$ & R@1$\uparrow$ & DCG$\uparrow$ \\
      \midrule
            \rowcolor{gray!17} \multicolumn{5}{l}{\textbf{ASGM temperature}} \\
      \noalign{\vskip 1pt}
      \rowcolor{lightblue}
      \textbf{$\tau$ = 0.7} & \textbf{34.75} & \textbf{61.24} & \textbf{57.66} & \textbf{77.21} \\
      $\tau$ = 1.0                        & 32.97 & 59.84 & 55.70 & 75.81 \\
      $\tau$ = 2.0                        & 30.15 & 56.27 & 53.16 & 73.00 \\
      \midrule
      Random selection                   & 29.66 & 55.81 & 52.22 & 72.58 \\
      \bottomrule
    \end{tabularx}
    \captionsetup{skip=2pt, belowskip=-18pt}
    \captionof{table}{
      \textbf{ASGM temperature.} Default highlighted ($\tau{=}0.7$). Larger $\tau$ softens the selection toward uniform, approaching random selection.}
    \label{tab:ablation_masking}
    \end{threeparttable}
\end{table}

\noindent\textbf{Selection temperature.}
The temperature $\tau$ controls how sharply ASGM concentrates on the dominant
articulator. As \cref{tab:ablation_masking} shows, the sharp $\tau{=}0.7$ is best,
reaching 34.75 R@1 on WLASL and 57.66 on ASL-Citizen. Raising $\tau$ flattens the
selection toward uniform and steadily lowers retrieval to 32.97 and 55.70 at
$\tau{=}1.0$ and 30.15 and 53.16 at $\tau{=}2.0$, approaching the random-selection
baseline of 29.66 and 52.22. Sharp, saliency-focused selection is what makes ASGM more effective.

\noindent\textbf{Articulator selection strategy.}
As \cref{tab:abl_selection} shows, replacing ASGM's per-clip choice with fixed or
heuristic selection rules lowers retrieval performance. A fixed choice forces the masked
objectives onto the same articulator every clip, so the model learns to model that
one and largely ignores the rest, which limits the representation and drives the
drop. On the hand-centric WLASL,
fixing selection to the face is worst at 24.78, below every hand-based rule, whereas
on the non-manual NMFs-CSL the face and body are the strongest fixed choices at 61.63
and 59.23, well above the 56.10 of restricting selection to the right hand. ASGM
instead adapts the choice to each input, selecting whichever articulator carries the
current sign so that, across training, the model learns all of them together with
their supporting context.

\begin{figure}[t]
    \centering
    \begin{subfigure}[b]{0.49\columnwidth}
        \includegraphics[width=\linewidth]{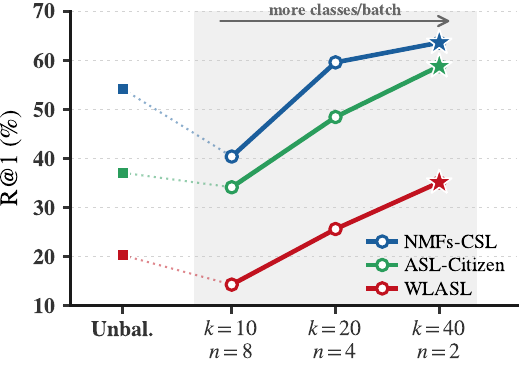}
        \caption{Batch composition}
        \label{fig:abl_batch}
    \end{subfigure}
    \begin{subfigure}[b]{0.49\columnwidth}
        \includegraphics[width=\linewidth]{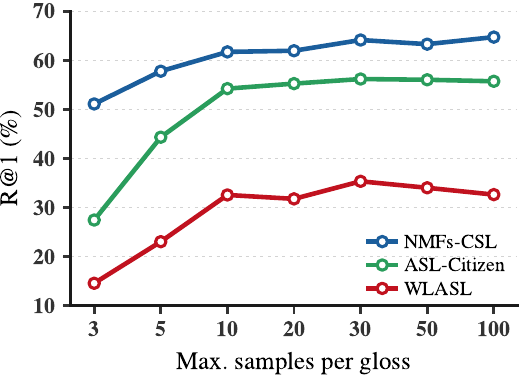}
        \caption{Data cap}
        \label{fig:abl_cap}
    \end{subfigure}
    \captionsetup{skip=2pt, belowskip=-18pt}
    \caption{\textbf{Batch composition and per-sign sample cap.} (a) More classes per batch (larger $k$) raises R@1
and beats the unbalanced baseline on all three datasets. (b) Retrieval saturates
by a cap of about $10$ clips per sign and peaks near $30$.}
\end{figure}

\noindent\textbf{Batch composition.}
Batch sampling strongly shapes the representation. Trading clips per class for more
classes at a fixed batch of $80$ steadily improves retrieval, with WLASL R@$1$ rising
from $14.28$ at $k{=}10,n{=}8$ to $34.8$ at $k{=}40,n{=}2$ and ASL-Citizen from
$34.12$ to $57.7$, a monotonic trend that holds on all three benchmarks
(\cref{fig:abl_batch}). Removing class balancing hurts on every dataset, consistent
with the contrastive principle that many distinct in-batch negatives matter more than
repeated views of a few classes.

\noindent\textbf{Sample cap per gloss.}
We vary the maximum number of clips each gloss contributes to training. Retrieval
improves sharply as the cap grows from 3 to 10, with WLASL R@1 rising from 14.56 to
32.56, ASL-Citizen from 27.44 to 54.23, and NMFs-CSL from 51.11 to 61.72. As
\cref{fig:abl_cap} shows, the curves then flatten and plateau at our full-model
performance of 34.75 on WLASL, 57.66 on ASL-Citizen, and 65.20 on NMFs-CSL, after
which a larger cap brings no further gain. The trend is consistent across datasets.

\begin{table}[t]
\centering
\footnotesize
\setlength{\tabcolsep}{5pt}
\renewcommand{\arraystretch}{1.1}
\begin{tabularx}{\columnwidth}{X r c r c}
\toprule
\multirow{2}{*}{\textbf{Articulator selection}}
 & \multicolumn{2}{c}{\textbf{WLASL}} & \multicolumn{2}{c}{\textbf{NMFs-CSL}} \\
\cmidrule(lr){2-3}\cmidrule(lr){4-5}
 & R@1$\uparrow$ & $\Delta$ & R@1$\uparrow$ & $\Delta$ \\
\midrule
Always body                    & 25.52 & \cellcolor{red!46}$-$9.23 & 59.23 & \cellcolor{red!30}$-$5.97 \\
Always face                    & 24.78 & \cellcolor{red!50}$-$9.97 & 61.63 & \cellcolor{red!18}$-$3.57 \\
Always right hand              & 26.40 & \cellcolor{red!42}$-$8.35 & 56.10 & \cellcolor{red!46}$-$9.10 \\
Always dominant hand$^\dagger$ & 31.65 & \cellcolor{red!16}$-$3.10 & 58.80 & \cellcolor{red!32}$-$6.40 \\
Uniform over hands (LH/RH)     & 30.41 & \cellcolor{red!22}$-$4.34 & 58.23 & \cellcolor{red!35}$-$6.97 \\
Uniform over all four          & 29.66 & \cellcolor{red!25}$-$5.09 & 62.78 & \cellcolor{red!12}$-$2.42 \\
\midrule
\rowcolor{lightblue}
\textbf{ASGM (ours)}           & \bfseries 34.75 & -- & \bfseries 65.20 & -- \\
\bottomrule
\end{tabularx}
\captionsetup{skip=2pt, belowskip=-8pt}
\caption{\textbf{Articulator selection strategy.} R@1 on WLASL and NMFs-CSL for
fixed and heuristic selection rules versus \textbf{ASGM}. $\Delta$ is the R@1 change
relative to ASGM, and darker red marks a larger drop.
$^\dagger$Per clip, chosen by confidence.}
\label{tab:abl_selection}
\end{table}

\vspace{-0.5em}

\section{Conclusion}
\label{sec:conclusion}
We introduced \textbf{SignSeek}, a pose-based pretraining 
framework that treats sign
dictionary retrieval as a primary objective. Beyond a contrastive objective, it adds
two articulator-aware masking objectives, guided by a saliency module that identifies
the dominant articulator, that recognise a sign through that articulator and recover
it from the 
context, encoding the articulatory structure of sign language
into the representation. 
\textbf{SignSeek} sets SOTA cross-corpus retrieval on three benchmarks, while generalizing to unseen BSL and transferring to sign recognition and subtitle alignment.

\vspace{-0.5em}
\section*{Acknowledgements}
{\sloppy
This work was supported by EPSRC grant APP24554 (SignGPT-EP/Z535370/1), EPSRC grant APP78083 (UMCS UKRI3927) and through funding from Google.org via the AI for Global Goals scheme.
The authors acknowledge the use of Isambard-AI National AI Research Resource (AIRR) funded by UK DSIT via UKRI and STFC [ST/AIRR/I-A-I/1023].\par}

{
    \small
    \bibliographystyle{ieeenat_fullname}
    \bibliography{main}
}

\clearpage
\pagenumbering{arabic}      
\setcounter{page}{1}
\setcounter{section}{0}
\setcounter{figure}{0}
\setcounter{table}{0}
\setcounter{equation}{0}
\renewcommand{\thesection}{\Alph{section}}
\addtocontents{toc}{\protect\setcounter{tocdepth}{3}}

\maketitlesupplementary
This document provides additional technical details that complement the main paper. We describe the architecture of our sign language backbone (\cref{supp:backbone}), the full pretraining and downstream training setup (\cref{supp:implementation}), additional ablation studies (\cref{supp:ablations}), the sign dictionary retrieval benchmarks together with qualitative results (\cref{supp:retrieval}), and the sign-subtitle alignment pipeline with further qualitative examples (\cref{sec:supp-ssa}).
{
    \hypersetup{linkcolor=black}
    \makeatletter\@starttoc{toc}\makeatother
}

\section{Architecture Details}
\label{supp:backbone}
\begin{figure}[t]
    \centering
    \includegraphics[width=1\linewidth]{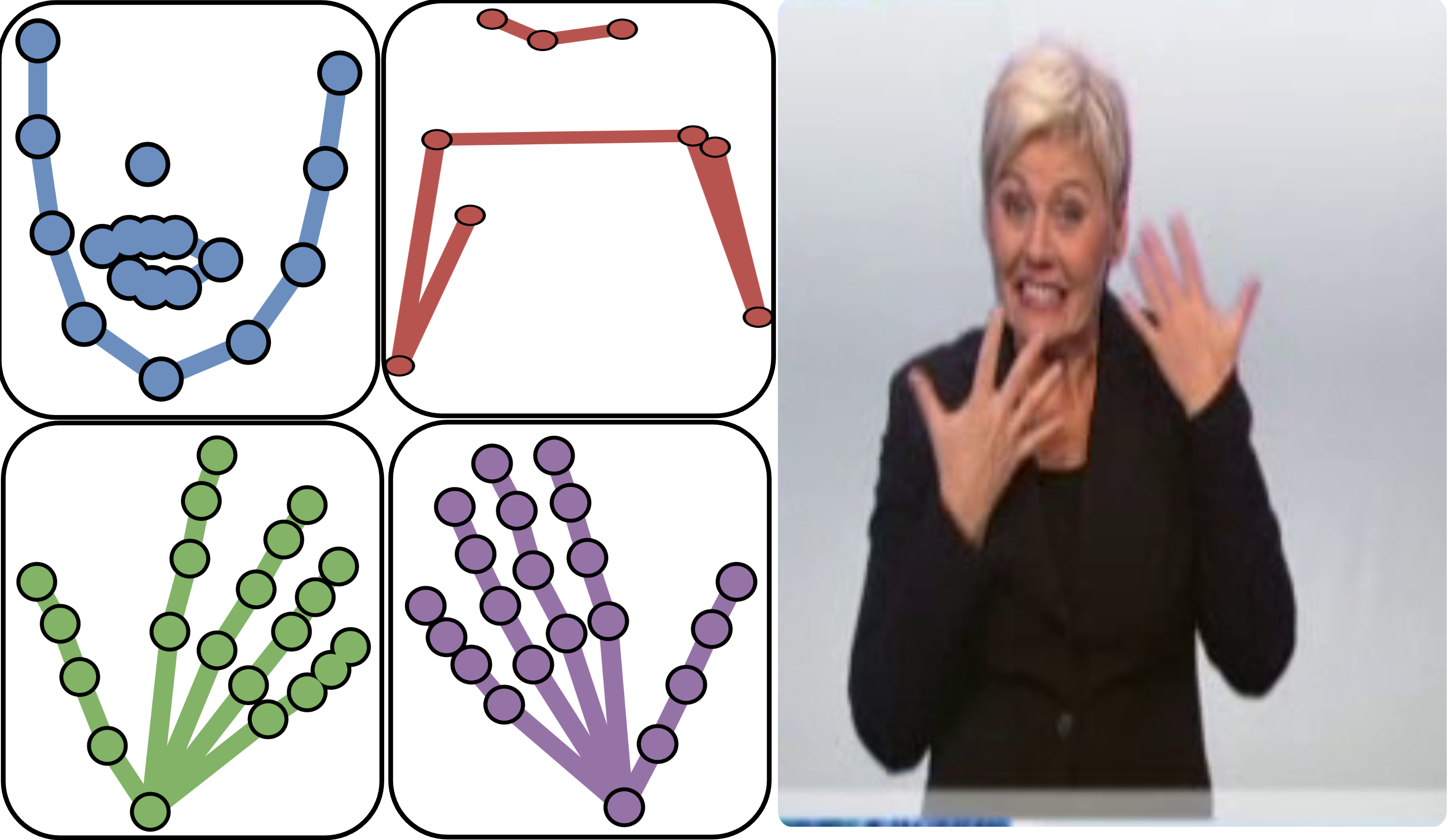}
    \caption{\textbf{Extracted articulator keypoints.} Example frames showing the
    RTMPose-x whole-body pose partitioned into the four articulators, body, face,
    left hand, and right hand, that form the input streams to our backbone.}
    \label{fig:keypoints}
\end{figure}
We provide details on our sign language backbone described in
\cref{sec:model} of the main paper. This section first explains our pose
extraction pipeline and regional partitioning (\cref{supp:pose_extraction}), then
the architectural specifications of the articulatory graph encoder and temporal
fusion module (\cref{supp:agcn_details}).

\subsection{Pose Extraction and Region Partitioning}
\label{supp:pose_extraction}
\noindent\textbf{Whole-Body Keypoint Extraction.}
We utilise RTMPose-x~\cite{jiang2023rtmpose} from the MMPose framework to extract
whole-body pose from sign language videos. RTMPose-x provides 133 keypoints
covering the entire body, including fine-grained hand and facial landmarks
essential for sign language understanding.

\noindent\textbf{Anatomical Region Partitioning.}
As described in \cref{sec:input} of the main paper, we partition the extracted
keypoints into four articulators, as illustrated in \cref{fig:keypoints}. The
specific keypoint indices for each region are as follows:

\begin{itemize}
    \item \textbf{Body} indices $\{1, 4\text{--}11\}$ ($J_{\mathrm{B}} = 9$ keypoints)
    \item \textbf{Left Hand} indices $\{92\text{--}112\}$ ($J_{\mathrm{LH}} = 21$ keypoints)
    \item \textbf{Right Hand} indices $\{113\text{--}133\}$ ($J_{\mathrm{RH}} = 21$ keypoints)
    \item \textbf{Face} indices $\{24, 26, 28, 30, 32, 34, \\ 36, 38, 40, 54, 84\text{--}91\}$ ($J_{\mathrm{F}} = 18$ keypoints)
\end{itemize}

\noindent\textbf{Root-Relative Normalisation.}
To ensure translation invariance and focus on relative motion, we apply
root-relative normalisation to the hands and face. We designate keypoint indices
92, 113, and 54 as root nodes for the left hand, right hand, and face,
respectively. Each keypoint in these regions is normalised relative to its root,
\begin{equation}
\mathcal{P}_{r}^{\text{norm}}(j) = \mathcal{P}_{r}(j) - \mathcal{P}_{r}(\text{root}), \quad j \in \mathcal{J}_r,
\end{equation}
where $\mathcal{J}_r$ denotes the keypoint indices of articulator $r$. The body
region is not normalised, preserving the absolute spatial position that captures
global body orientation.

\subsection{Input Representation to the AGCN Encoders}
The 2D keypoint sequences $\mathcal{P}_{r} \in \mathbb{R}^{T \times J_{r} \times 2}$
are extended with the RTMPose confidence scores before entering the AGCN encoders.
Each keypoint $(x, y)$ becomes $(x, y, c)$, where $c \in [0, 1]$ is the detection
confidence, so the input has shape $\mathbb{R}^{T \times J_{r} \times 3}$. Each
encoder maps this 3-channel input to $C_{\text{in}}=64$ channels with a per-joint
linear projection before the adaptive graph convolutions. The confidence channel
informs the network about keypoint reliability, which is valuable for handling
occlusion and detection uncertainty in sign language video.

\begin{table}[t]
\centering
\small
\resizebox{\columnwidth}{!}{
\begin{tabular}{l|l|c|c}
\hline
\textbf{Stage} & \textbf{Layer Type} & \textbf{Channels} & \textbf{Temporal Kernel} \\
\hline
Input Projection & Linear            & $3 \rightarrow 64$    & -- \\
\hline
\rowcolor{gray!30}
\multicolumn{4}{c}{\textbf{Adaptive Graph Blocks}} \\
\hline
Block 1 & AGConv + TCN + SE & $64 \rightarrow 128$   & 7 \\
Block 2 & AGConv + TCN + SE & $128 \rightarrow 128$  & 7 \\
Block 3 & AGConv + TCN + SE & $128 \rightarrow 192$  & 7 \\
Block 4 & AGConv + TCN + SE & $192 \rightarrow 384$  & 7 \\
\hline
Joint Aggregation & Flatten + Linear & $J\!\cdot\!384 \rightarrow 384$ & -- \\
\hline
\multicolumn{4}{l}{\footnotesize AGConv: adaptive graph conv (3 subsets, learned adjacency).} \\
\multicolumn{4}{l}{\footnotesize TCN: temporal conv. SE: squeeze-and-excitation (reduction 16).} \\
\multicolumn{4}{l}{\footnotesize All blocks include residual connections, BN, ReLU, and Dropout.} \\
\multicolumn{4}{l}{\footnotesize Output channels $C = 6\,C_{\text{in}} = 384$.} \\
\end{tabular}}
\caption{\textbf{Per-articulator AGCN encoder}.}
\label{tab:agcn_detailed}
\end{table}

\subsection{Articulatory Backbone Details}
\label{supp:agcn_details}
The backbone encodes each articulator with an adaptive graph convolutional (AGCN)
encoder, fuses the four streams with a Conformer, and pools the result into a
single clip embedding. We detail each component below.

\noindent\textbf{Articulatory Graph Encoder.}
Each articulator stream $X^a \in \mathbb{R}^{T \times J_a \times 3}$ is encoded
independently. A per-joint linear projection lifts each keypoint to
$C_{\text{in}}=64$ channels, followed by four adaptive graph convolutional blocks
that widen the features to $C = 6\,C_{\text{in}} = 384$. Each block couples an
adaptive spatial graph convolution, whose adjacency is learned rather than fixed
to the skeleton, with a temporal convolution and a squeeze-and-excitation block
that recalibrates channels~\cite{shi2019two}. Instead of pooling over joints, the per-frame joint
map is flattened and projected back to $C$ channels by a learned aggregation, so
the encoder learns a per-joint weighting rather than treating all joints equally.
The body and face each use a dedicated encoder, while the two hands share a single
encoder with the left hand mirrored by an $x$-axis flip $\Phi_x$.
\cref{tab:agcn_detailed} gives the per-articulator layer stack.

\noindent\textbf{Temporal Fusion Encoder.}
The four articulator streams are concatenated into a $T \times 1536$ sequence and
linearly projected to the model dimension $d_m = 256$, then passed to a Conformer
encoder. Each Conformer block interleaves multi-head self-attention with a
depthwise convolution module, wrapped by two half-step feed-forward layers in the
Macaron configuration~\cite{gulati2020conformer}, and sinusoidal positional
encodings are added before the first block. This captures long-range temporal
structure and yields a contextualised sequence
$\mathbf{s} \in \mathbb{R}^{T \times d_m}$. \cref{tab:conformer} lists the
configuration.

\begin{table}[t]
\centering
\footnotesize
\begin{tabular*}{\columnwidth}{@{\extracolsep{\fill}}ll}
\hline
\textbf{Component} & \textbf{Value} \\
\hline
Model dimension $d_m$        & 256 \\
Number of blocks             & 2 \\
Attention heads              & 4 \\
Feed-forward expansion       & $2\times$ ($256 \rightarrow 512$) \\
Convolution kernel size      & 5 \\
Positional encoding          & Sinusoidal \\
Dropout                      & 0.1 \\
\hline
\end{tabular*}
\caption{\textbf{Conformer temporal fusion encoder}.}
\label{tab:conformer}
\end{table}

\noindent\textbf{Attentive Pooling and Projection.}
The contextualised sequence is reduced to a single clip embedding by an attentive
pooling head that reuses the Conformer self-attention to score each frame and
takes a weighted average over time, mapping the $T \times 256$ sequence to a
single $256$-dimensional vector. A two-layer MLP then projects this vector to the
final $128$-dimensional embedding $z$, which we L2-normalise and use for the
contrastive objectives and, at inference, as the retrieval embedding. The complete end-to-end backbone is summarised in \cref{tab:agcn_architecture}.

\noindent\textbf{Predictor.}
The masked prediction objective (MAP, \cref{sec:objectives} of the main paper)
relies on a lightweight predictor $g$ that maps the Conformer output sequence to
the per-frame target representation. It is a two-layer Transformer decoder in
which learnable positional embeddings for all frame positions act as queries and
the Conformer output $\mathbf{s} \in \mathbb{R}^{T \times 256}$ serves as memory,
so every predicted frame can attend to the full temporal context. Each decoder
layer uses 4 attention heads, a feed-forward dimension of $1024$, and pre-layer
normalisation. A final layer normalisation and a linear projection map each frame
to the $384$-dimensional articulator target. The predictor is used only during
training and is discarded at inference.

\begin{table}[]
\centering
\small
\resizebox{\columnwidth}{!}{
\begin{tabular}{c|c|c|c|c}
\hline
\textbf{Stream} & \textbf{Component} & \textbf{Input} & \textbf{Output} & \textbf{Graph} \\
\hline
\multirow{2}{*}{\makecell{\textbf{Body} \\ $(E_{\mathrm{B}})$}}
 & AGCN Encoder     & $T \times J_{\mathrm{B}} \times 3$ & $T \times 384$ & Adaptive \\
 & Joint Aggregation & $T \times J_{\mathrm{B}} \times 384$ & $T \times 384$ & -- \\
\hline
\multirow{2}{*}{\makecell{\textbf{Face} \\ $(E_{\mathrm{F}})$}}
 & AGCN Encoder     & $T \times J_{\mathrm{F}} \times 3$ & $T \times 384$ & Adaptive \\
 & Joint Aggregation & $T \times J_{\mathrm{F}} \times 384$ & $T \times 384$ & -- \\
\hline
\multirow{2}{*}{\makecell{\textbf{Right Hand} \\ $(E_{\mathrm{H}})$}}
 & AGCN Encoder (shared) & $T \times J_{\mathrm{RH}} \times 3$ & $T \times 384$ & Adaptive \\
 & Joint Aggregation & $T \times J_{\mathrm{RH}} \times 384$ & $T \times 384$ & -- \\
\hline
\multirow{2}{*}{\makecell{\textbf{Left Hand} \\ $(E_{\mathrm{H}})$}}
 & AGCN Encoder (shared, $\Phi_x$) & $T \times J_{\mathrm{LH}} \times 3$ & $T \times 384$ & Adaptive \\
 & Joint Aggregation & $T \times J_{\mathrm{LH}} \times 384$ & $T \times 384$ & -- \\
\hline
\multicolumn{2}{l|}{\textbf{Fusion (Concat)}} & \multicolumn{3}{c}{$T \times 1536$} \\ \hline
\multicolumn{2}{l|}{\textbf{Feature Projection}} & $T \times 1536$ & $T \times 256$ & -- \\ \hline
\multicolumn{2}{l|}{\textbf{Conformer} (2 blocks, 4 heads)} & $T \times 256$ & $T \times 256$ & -- \\ \hline
\multicolumn{2}{l|}{\textbf{Attentive Pooling}} & $T \times 256$ & $256$ & -- \\ \hline
\multicolumn{2}{l|}{\textbf{Projection Head} (MLP)} & $256$ & $128$ & -- \\ \hline
\multicolumn{2}{l|}{\textbf{Number of Parameters}} & \multicolumn{3}{c}{15.7\,M} \\ \hline
\end{tabular}
}
\caption{\textbf{Sign Language Backbone Architecture.} $T$: temporal length, $J_{\mathrm{B}}/J_{\mathrm{F}}/J_{\mathrm{RH}}/J_{\mathrm{LH}}$: body/face/right/left-hand keypoints. The two hands share one encoder, with the left hand mirrored by an $x$-axis flip $\Phi_x$.}
\label{tab:agcn_architecture}
\end{table}

\section{Implementation Details}
\label{supp:implementation}
\subsection{Pretraining}
We provide the full pretraining recipe for \textbf{SignSeek}. All hyperparameters
are summarised in \cref{tab:pretrain_hparams}.

\noindent\textbf{Optimisation.}
We pretrain the backbone with the AdamW optimiser~\cite{adamw} at a base learning rate of
$3\times10^{-4}$ and a weight decay of $0.01$, using a cosine learning-rate
schedule with a linear warm-up over the first 3{,}000 steps. Training runs for
30{,}000 steps in total, and we select the checkpoint with the lowest validation loss.
Optimisation uses \texttt{bf16} mixed precision through the DeepSpeed framework.
Pretraining is carried out on a single NVIDIA RTX PRO 6000, while all retrieval and downstream evaluations are run on a single
NVIDIA RTX 3090.

\noindent\textbf{Batch construction.}
We use class-balanced sampling to guarantee positive pairs for the contrastive
objectives. Each batch draws 40 glosses and 2 clips per gloss, and the two clips
of a gloss serve as the query and reference views used by the GAC and MAC
objectives.

\subsection{Downstream Recognition}
For isolated sign recognition, we fine-tune the pretrained backbone with a linear
classification head. During training we augment each clip in keypoint space,
sampling independently a random in-plane rotation in $[-10^{\circ}, 10^{\circ}]$,
an isotropic scale in $[0.9, 1.1]$, and a translation in $[-0.1, 0.1]$ of the
normalised coordinate range, and no augmentation is applied at inference. We train
with a cross-entropy loss with label smoothing of $0.1$ and no feature masking,
optimising with AdamW~\cite{adamw} at a maximum learning rate of
$1\times10^{-4}$ and a batch size of 32 for 100 epochs, using 1{,}000 warm-up
iterations followed by cosine annealing of the learning rate. A layer-wise
learning-rate decay of $0.85$ is applied across the backbone, so earlier layers
adapt more slowly than the classification head.

\begin{table}[t]
\centering
\footnotesize
\begin{tabular*}{\columnwidth}{@{\extracolsep{\fill}}ll}
\hline
\textbf{Hyperparameter} & \textbf{Value} \\
\hline
Optimiser                       & AdamW \\
Base learning rate              & $3\times10^{-4}$ \\
Weight decay                    & 0.01 \\
Optimiser betas                 & $(0.9, 0.98)$ \\
LR schedule                     & Cosine \\
Warm-up steps                  & 3,000 \\
Total steps                    & 30,000 \\
Mixed precision                 & bf16 (DeepSpeed) \\
\hline
Glosses per batch               & 40 \\
Clips per gloss                 & 2 \\
Contrastive temperature         & 0.2 \\
ASGM temperature $\tau$         & 0.7 \\
Loss weights $\lambda_{\text{GAC}}/\lambda_{\text{MAC}}/\lambda_{\text{MAP}}$ & $1.0/1.0/1.0$ \\
\hline
\end{tabular*}
\caption{\textbf{Pretraining hyperparameters.}}
\label{tab:pretrain_hparams}
\end{table}

\section{Additional Ablation Studies}
\label{supp:ablations}
\newcolumntype{Y}{>{\centering\arraybackslash}X}
\definecolor{lightgold}{RGB}{253,236,160}

\subsection{Training Objectives}
\cref{tab:abl_objectives} extends the objective ablation of the main paper with the
two single masked objectives, MAC only and MAP only. The shared configurations are
analysed in the main paper (\cref{tab:ablation_retrieval}), which shows that GAC is
a strong baseline and that adding MAC and MAP on top of it is complementary, so we
focus here on the two new rows. Used on their own, without the GAC anchor, neither
masked objective improves on the plain GAC baseline, with MAC only reaching $27.09$
R@$1$ on WLASL and $48.06$ on ASL-Citizen and MAP only $25.65$ and $47.70$, against
$27.25$ and $48.40$ for GAC, and both far below the full model. When either masked
objective is trained on its own without the GAC anchor, the backbone is not
discriminative enough for ASGM to tell which articulator carries the sign, so the
saliency-guided masking selects poorly and the objective cannot improve on the
contrastive baseline.

\begin{table}[t]
    \centering
    \footnotesize
    \setlength{\tabcolsep}{4pt}
    \begin{threeparttable}
    \begin{tabularx}{\columnwidth}{ccc YYYY}
      \toprule
      & & & \multicolumn{2}{c}{\textbf{WLASL}} & \multicolumn{2}{c}{\textbf{ASL-Citizen}} \\
      \cmidrule(lr){4-5}\cmidrule(lr){6-7}
      \textbf{GAC} & \textbf{MAC} & \textbf{MAP} & R@1$\uparrow$ & DCG$\uparrow$ & R@1$\uparrow$ & DCG$\uparrow$ \\
      \midrule
      \cmark & \xmark & \xmark & 27.25 & 54.02 & 48.40 & 70.02 \\
      \cmark & \cmark & \xmark & 31.27 & 58.55 & 54.15 & 75.02 \\
      \cmark & \xmark & \cmark & 32.33 & 60.16 & 55.93 & 75.77 \\
      \xmark & \cmark & \cmark & 32.45 & 60.20 & 55.12 & 75.05 \\
      \midrule
      \rowcolor{lightblue}
      \cmark & \cmark & \cmark & \textbf{34.75} & \textbf{61.24} & \textbf{57.66} & \textbf{77.21} \\
      \midrule
      \rowcolor{lightgold}
      \xmark & \cmark & \xmark & 27.09 & 53.25 & 48.06 & 69.15 \\
      \rowcolor{lightgold}
      \xmark & \xmark & \cmark &  25.65 & 50.75 & 47.70 & 68.60 \\
      \bottomrule
    \end{tabularx}
\captionof{table}{\textbf{Objective ablation.} Each row toggles GAC, MAC, and MAP.
The full model with all three (blue) is \textbf{SignSeek}. The two gold rows,
MAC only and MAP only, are additional configurations added in this supplement.}
    \label{tab:abl_objectives}
    \end{threeparttable}
\end{table}

\subsection{Loss Weighting}
Our joint objective weights the three terms equally with
$\lambda_{\text{GAC}}=\lambda_{\text{MAC}}=\lambda_{\text{MAP}}=1$. Here we vary
the weights to assess sensitivity, keeping every other setting fixed.
\cref{tab:abl_loss_weight} reports the results. Equal weighting gives the best R@1
and DCG on both datasets, and no reweighting improves over it, so we keep it
throughout the paper.

\begin{table}[t]
\centering
\footnotesize
\setlength{\tabcolsep}{4pt}
\begin{threeparttable}
\begin{tabularx}{\columnwidth}{ccc YYYY}
\toprule
& & & \multicolumn{2}{c}{\textbf{WLASL}} & \multicolumn{2}{c}{\textbf{ASL-Citizen}} \\
\cmidrule(lr){4-5}\cmidrule(lr){6-7}
$\lambda_{\text{GAC}}$ & $\lambda_{\text{MAC}}$ & $\lambda_{\text{MAP}}$
& R@1$\uparrow$ & DCG$\uparrow$ & R@1$\uparrow$ & DCG$\uparrow$ \\
\midrule
\rowcolor{lightblue}
1 & 1 & 1 & \textbf{34.75} & \textbf{61.24} & \textbf{57.66} & \textbf{77.21} \\
0.5 & 1 & 1 & 34.04 & 61.20 & 56.80 & 76.50 \\
1 & 0.5 & 1 & 33.90 & 60.90 & 56.70 & 75.90 \\
1 & 1 & 0.5 & 34.60 & 61.22 & 56.63 & 76.85 \\
0.5 & 0.5 & 1 & 32.54 & 60.75 & 55.25 & 75.15 \\
0.5 & 1 & 0.5 & 33.94 & 61.05 & 57.22 & 77.18 \\
1 & 0.5 & 0.5 & 34.35 & 61.00 & 56.60 & 76.90 \\
\bottomrule
\end{tabularx}
\captionof{table}{\textbf{Loss-weighting ablation.} Effect of varying the objective
weights $\lambda_{\text{GAC}}$, $\lambda_{\text{MAC}}$, and $\lambda_{\text{MAP}}$.
The equal-weight setting (highlighted) is used in \textbf{SignSeek}.}
\label{tab:abl_loss_weight}
\end{threeparttable}
\end{table}

\section{Sign Dictionary Retrieval}
\label{supp:retrieval}

\subsection{Evaluation Datasets}
We evaluate sign dictionary retrieval on three isolated-sign benchmarks that are
held out from our pretraining corpus, WLASL~\cite{wlasl},
ASL-Citizen~\cite{desai2023asl}, and NMFs-CSL~\cite{Hu_2021}. None of the three is
seen during pretraining, so every result measures cross-corpus generalisation of
the frozen encoder without any downstream fine-tuning. Their statistics are listed
in \cref{tab:retrieval-datasets}. Following our protocol, we treat each test split
as the set of queries and the corresponding training split as the retrieval
gallery, ranking every gallery clip by cosine similarity to the query embedding.
The benchmarks span a wide range of vocabulary sizes, from 1{,}067 glosses in
NMFs-CSL to 2{,}731 in ASL-Citizen, and gallery sizes from roughly 14K to 40K
clips, so retrieval difficulty grows with both the number of distractors and the
vocabulary.

\begin{table}[t]
\centering
\footnotesize
\begin{tabular*}{\columnwidth}{@{\extracolsep{\fill}}lcrrr}
\hline
\textbf{Dataset} & \textbf{Lang.} & \textbf{Glosses} & \textbf{Gallery} & \textbf{Query} \\
\hline
WLASL~\cite{wlasl}          & ASL & 2{,}000 & 14{,}123 &  2{,}878 \\
ASL-Citizen~\cite{desai2023asl} & ASL & 2{,}731 & 40{,}154 & 32{,}941 \\
NMFs-CSL~\cite{Hu_2021}     & CSL & 1{,}067 & 25{,}608 &  6{,}402 \\
\hline
\end{tabular*}
\caption{\textbf{Sign dictionary retrieval benchmarks}. Gallery and query counts are the
training and test splits, respectively.}
\label{tab:retrieval-datasets}
\end{table}

\subsection{BSL Evaluation Set}
To probe cross-lingual transfer, we additionally evaluate on British Sign Language,
a language that is entirely held out from pretraining. All BSL sources, including
SignBank, are filtered out of the pretraining corpus, so the model has never seen
any BSL data before this evaluation. The set consists of $726$ query--gloss pairs
recorded by a native BSL signer, each annotated with its canonical gloss label. We
query these clips against BSL SignBank~\cite{signbank}, a publicly available
reference lexicon of British Sign Language that serves as the retrieval gallery,
which mirrors how a learner would look up an unfamiliar sign against a reference
dictionary. As in the other benchmarks, the frozen encoder embeds each query and
ranks the gallery entries by cosine similarity, and we report R@$k$ and mAP.
Because no BSL data is used at any stage of training, this setting measures purely
cross-lingual generalisation to a new sign language and vocabulary.

\subsection{Baselines}
Sign dictionary retrieval is only reported by SignRep~\cite{wong2025signrep}, so in
\cref{tab:retrieval} the pose-feature baselines, HieraMAE~\cite{ryali2023hiera},
and SignRep itself are quoted directly from the SignRep paper. The remaining
methods, I3D~\cite{Albanie2020bsl1k}, Video-Swin~\cite{prajwal2022weakly},
SignCLIP~\cite{jiang2024signclip}, and MASA~\cite{zhao2024masa}, do not publish
retrieval numbers, so we compute them ourselves. For these, we take the authors'
publicly released weights and run each model as a frozen feature extractor,
following their own recipe for extracting embeddings and without re-training or
fine-tuning. All methods are then evaluated under the same protocol as
\textbf{SignSeek}, using the same query and gallery splits, cosine-similarity
ranking, and DCG, R@1, and R@5 metrics, so the comparison is fair.

\subsection{Qualitative Results}
\begin{figure*}[tp]
    \centering
    \includegraphics[width=1\linewidth]{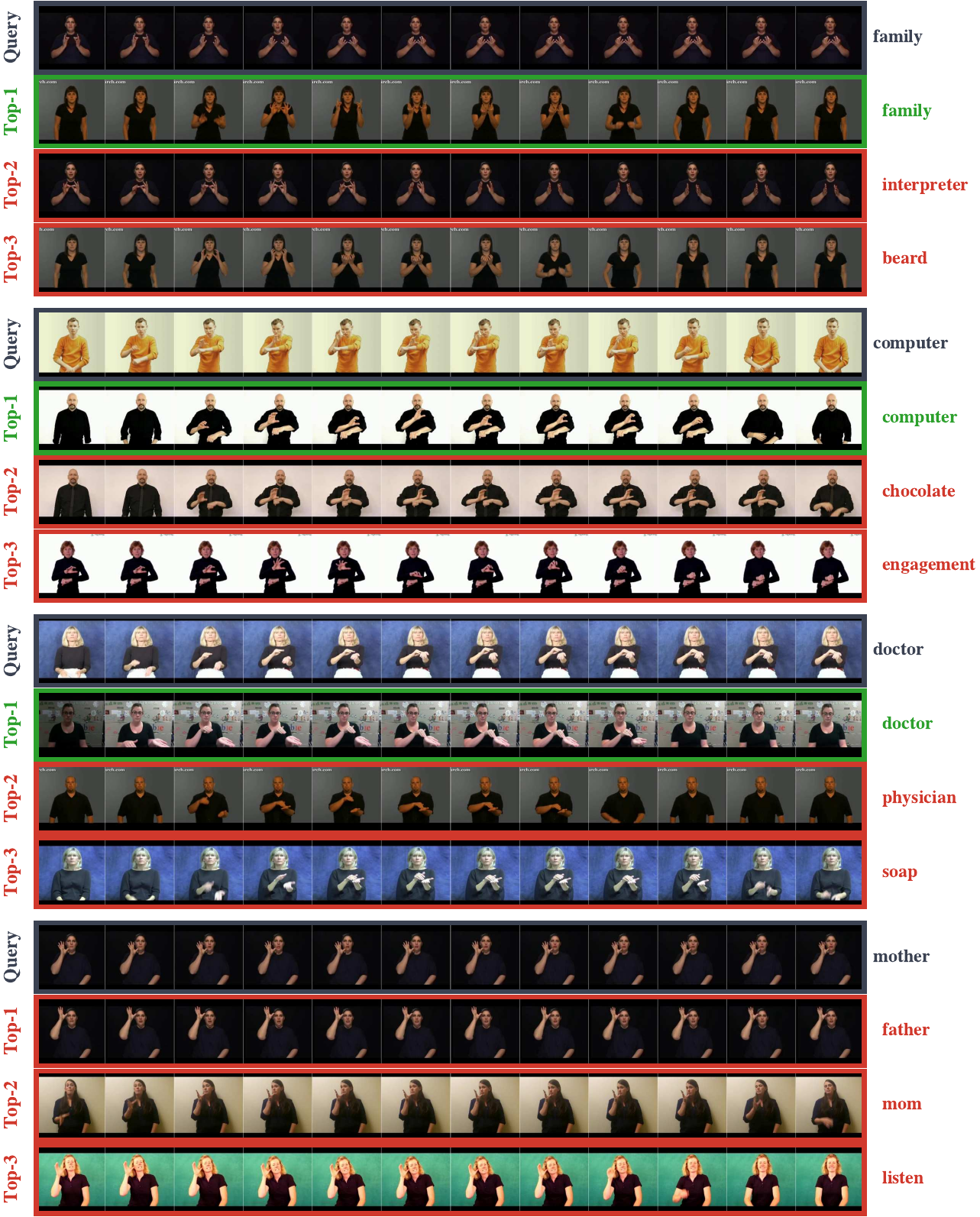}
    \caption{\textbf{Qualitative retrieval on WLASL.} For each query sign (top, from
    the test set), we show the top-3 distinct glosses retrieved by \textbf{SignSeek}
    from the training gallery, ordered by cosine similarity. Correct matches are
    outlined in green and incorrect ones in red.}
    \label{fig:qual_wlasl}
\end{figure*}

\begin{figure*}[tp]
    \centering
    \includegraphics[width=1\linewidth]{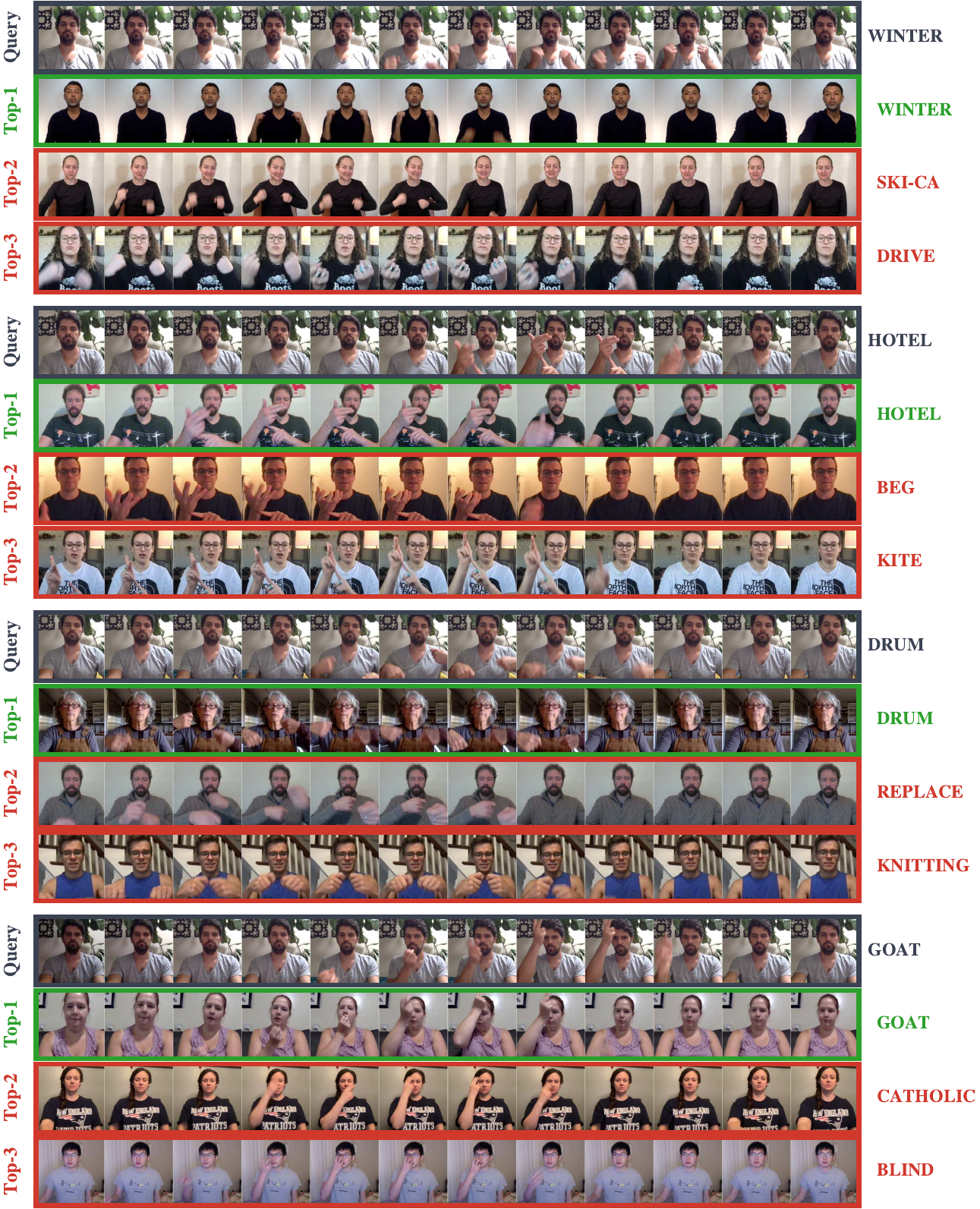}
    \caption{\textbf{Qualitative retrieval on ASL-Citizen.} For each query sign (top,
    from the test set), we show the top-3 distinct glosses retrieved by
    \textbf{SignSeek} from the training gallery, ordered by cosine similarity.
    Correct matches are outlined in green and incorrect ones in red.}
    \label{fig:qual_aslcitizen}
\end{figure*}

\begin{figure*}[tp]
    \centering
    \includegraphics[width=1\linewidth]{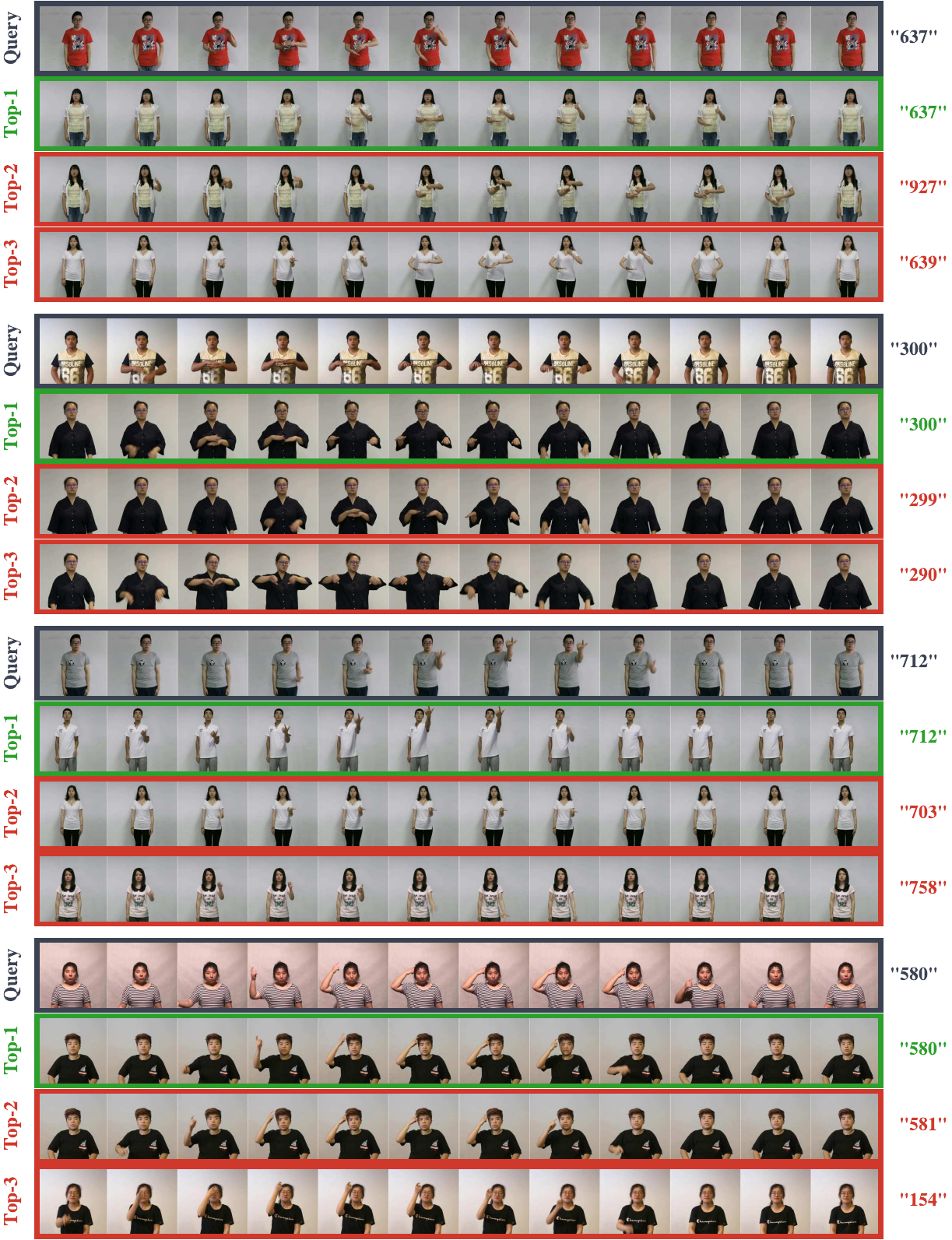}
    \caption{\textbf{Qualitative retrieval on NMFs-CSL.} For each query sign (top,
    from the test set), we show the top-3 distinct glosses retrieved by
    \textbf{SignSeek} from the training gallery, ordered by cosine similarity.
    Correct matches are outlined in green and incorrect ones in red. As gloss labels
    are not available for NMFs-CSL, signs are identified by their class IDs.}
    \label{fig:qual_nmfcsl}
\end{figure*}

\begin{figure*}[tp]
    \centering
    \includegraphics[width=\linewidth]{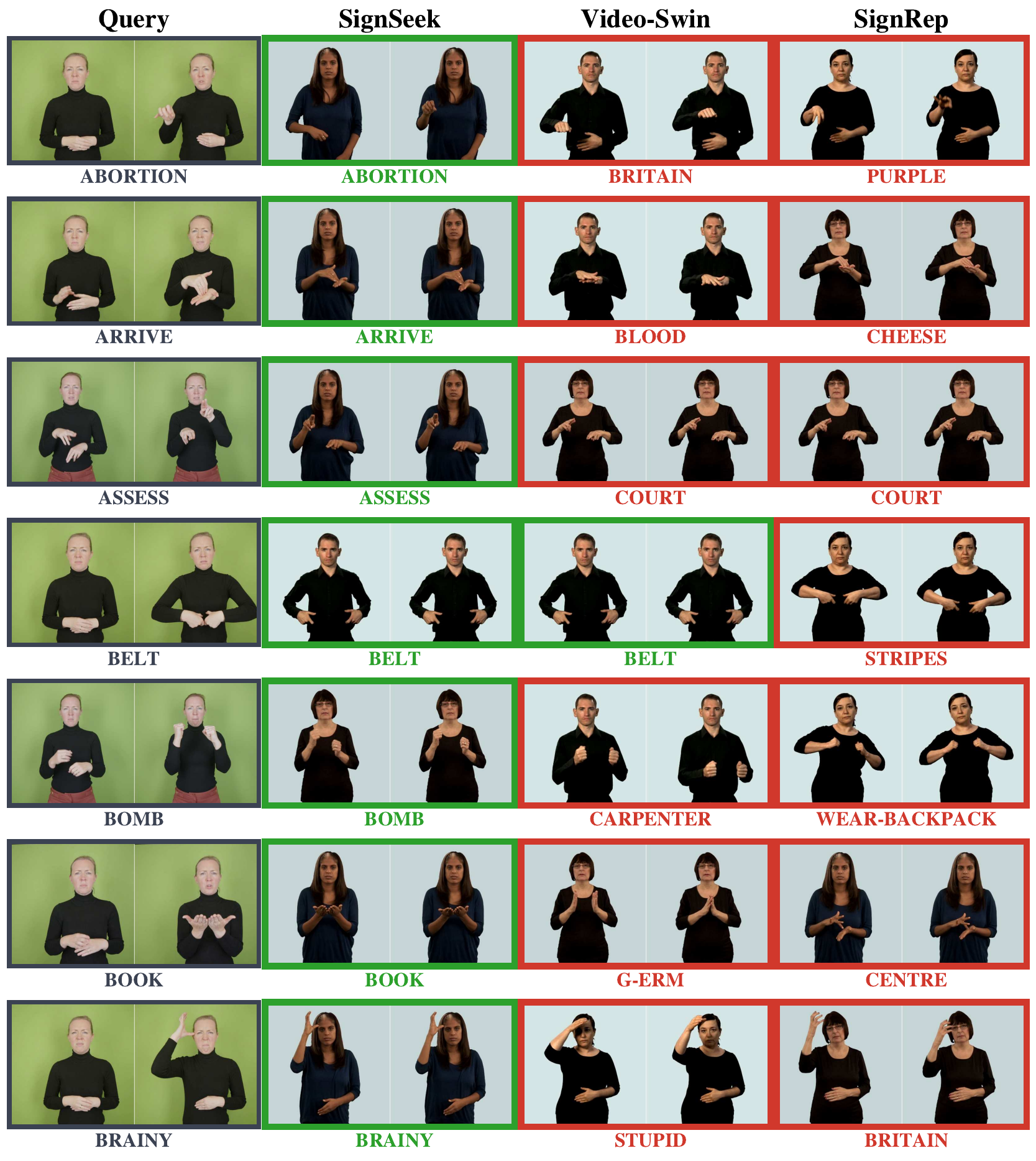}
    \caption{\textbf{Top-1 retrieval on the BSL evaluation set.} For each
    native-signer query (left, from the BSL evaluation set), we compare the single
    highest-ranked SignBank gloss retrieved by \textbf{SignSeek},
    Video-Swin~\cite{prajwal2022weakly}, and SignRep~\cite{wong2025signrep}, with all
    three applied as frozen encoders under the same protocol. Correct matches are
    outlined in green and incorrect ones in red. As no BSL data is seen during
    pretraining, these examples illustrate purely cross-lingual retrieval, where
    \textbf{SignSeek} recovers the correct gloss more often than the two prior
    methods.}
    \label{fig:qual_top1}
\end{figure*}

We visualise retrieval examples in
\cref{fig:qual_wlasl,fig:qual_aslcitizen,fig:qual_nmfcsl,fig:qual_top1}. For each
query, the frozen encoder embeds the clip and ranks the gallery by cosine
similarity. Since the gallery can contain multiple clips of the same sign, we keep
only the highest-ranked clip of each gloss, so every displayed result is a distinct
candidate sign rather than a repeated exemplar.
\cref{fig:qual_wlasl,fig:qual_aslcitizen,fig:qual_nmfcsl}
show the top-3 distinct glosses retrieved by \textbf{SignSeek} on WLASL,
ASL-Citizen, and NMFs-CSL, while \cref{fig:qual_top1} compares the top-1 retrieved
gloss of \textbf{SignSeek} against Video-Swin~\cite{prajwal2022weakly} and
SignRep~\cite{wong2025signrep} on our BSL evaluation set. A result is marked correct
(green) when its gloss matches the query gloss and incorrect (red) otherwise.


\clearpage

\section{Sign-Subtitle Alignment}
\label{sec:supp-ssa}

Sign-subtitle alignment (SSA) tightens loosely time-stamped subtitles onto the
frames where their signs are actually produced. We treat it as a downstream
application of the frozen \textbf{SignSeek} encoder $\Phi$, with no alignment-specific
training. Given a continuous signing video with $N$ subtitles whose SRT timings
$[\hat a_i,\hat b_i]$ are only coarse, we retrieve dictionary exemplars for the
words in each subtitle, match them against the video with $\Phi$, and decode a
monotonic assignment of subtitles to video segments. \cref{fig:ssa_qual1}
and \cref{fig:ssa_bobsl} show qualitative results on How2Sign~\cite{how2sign} and BOBSL~\cite{bobsl}.

\noindent\textbf{Pseudo-glossing.}
Subtitles are natural-language sentences rather than gloss sequences, so we first
reduce each subtitle $T_i$ to a small set of dictionary-matchable cues. We
lemmatise $T_i$ and keep content words (nouns, verbs, adjectives, adverbs,
numerals, and proper nouns), then drop personal, possessive, and reflexive
pronouns whose isolated dictionary exemplars are dominated by clip-level
artefacts and carry no useful localisation signal. Retaining the lemmas that are
present in the dictionary yields a pseudo-gloss set
$\mathcal{G}_i=\{g_{i,1},\dots,g_{i,L_i}\}$. Subtitles that contribute no
in-dictionary gloss keep their rough SRT timing unchanged.

\begin{table}[t]
  \centering
  \footnotesize
  \begin{tabular*}{\linewidth}{@{\extracolsep{\fill}}llrrr}
    \toprule
    Lang. & Source & Clips & Glosses & Ex./gloss \\
    \midrule
    \multirow{5}{*}{ASL}
      & ASL-Citizen~\cite{desai2023asl}    & 83{,}399  & 2{,}731 & 30.5 \\
      & SemLex~\cite{kezar2023sem}          & 41{,}597  & 3{,}097 & 13.4 \\
      & WLASL~\cite{wlasl}           & 18{,}039  & 1{,}900 &  9.5 \\
      & MSASL~\cite{joze2019msasl}           &  1{,}911  &   173   & 11.0 \\
    \cmidrule(l){2-5}
      & \textbf{Total} & \textbf{144{,}946} & \textbf{5{,}022}$^\dagger$ & \textbf{28.9} \\
    \midrule
    \multirow{3}{*}{BSL}
      & bsldict~\cite{signbsl_hello}        & 13{,}858  & 8{,}944 & 1.5 \\
      & SignBank~\cite{signbank}       &  3{,}453  & 2{,}033 & 1.7 \\
    \cmidrule(l){2-5}
      & \textbf{Total} & \textbf{17{,}311} & \textbf{9{,}534}$^\dagger$ & \textbf{1.8} \\
    \bottomrule
  \end{tabular*}
  \caption{\textbf{Dictionaries used for sign-subtitle alignment.} $^\dagger$Unique glosses, not the sum of per-source counts (glosses overlap across corpora). The ASL
  dictionary averages $\sim\!29$ exemplars per gloss against only $\sim\!1.8$
  for BSL.}
  \label{tab:ssa-dicts}
\end{table}

\noindent\textbf{Prototype banks and window similarity.}
For every gloss $g$ we collect its isolated exemplars from a language-matched
dictionary (ASL for How2Sign, BSL for BOBSL) and embed each with the frozen
encoder into a prototype set $P_g$. The two dictionaries differ sharply in
density. The ASL dictionary pools four isolated-sign corpora (ASL-Citizen,
SemLex, WLASL, and MSASL) into roughly $145$K clips over $5$K glosses, around
$29$ exemplars per gloss, whereas the BSL dictionary draws on two lexicon sources
(bsldict and SignBank) with only $17$K clips over $9.5$K glosses, around $1.8$
exemplars per gloss (\cref{tab:ssa-dicts}). The sparse BSL banks often provide a single
exemplar per gloss. The video is then tiled
into $M$ overlapping windows $\{w_j\}_{j=1}^{M}$ of length $W$ and stride
$\delta$, each embedded as $u_j=\Phi(w_j)$. We form a gloss-to-window similarity
matrix that records, for each subtitle, the strongest matching prototype at each
window,
\begin{equation}
S_{i,j}=\max_{g\in\mathcal{G}_i}\ \max_{p\in P_g}\ \cos\!\big(p,\,u_j\big).
\label{eq:supp-ssa-sim}
\end{equation}
The two-level maximum keeps the brightest single-gloss evidence per window, so a
subtitle is localised wherever any of its signs appears rather than requiring all
of them to co-occur.

\noindent\textbf{Monotonic decoding.}
Each subtitle is assigned a contiguous block of windows $[s_i,e_i]$, giving the
predicted span $[\tau^{-}_{s_i},\tau^{+}_{e_i}]$, where $\tau^{-}_j$ and
$\tau^{+}_j$ are the start and end times of window $j$. The cost of a block trades
agreement with the rough SRT timing against the retrieved similarity,
\begin{equation}
\begin{split}
\phi_i(s,e)={}&
w_{\text{d}}\big(|\hat a_i-\tau^{-}_{s}|+|\hat b_i-\tau^{+}_{e}|\big)\\
&+w_{\ell}\,\big|\Delta_{s,e}-\hat\Delta_i\big|
-w_{\text{s}}\!\sum_{j=s}^{e} S_{i,j},
\end{split}
\label{eq:supp-ssa-cost}
\end{equation}
where $\Delta_{s,e}=\tau^{+}_{e}-\tau^{-}_{s}$ is the block duration and
$\hat\Delta_i=\hat b_i-\hat a_i$ the rough duration. The first two terms anchor
the prediction near the SRT boundaries and discourage implausible durations,
while the third rewards blocks that cover high-similarity windows. Subtitles are
decoded left to right under a monotonic constraint that forbids overlap, so
subtitle $i{+}1$ begins strictly after subtitle $i$ ends. The optimal assignment
minimises the total cost and is recovered by dynamic programming recurrence,
\begin{equation}
D_i(e)=\min_{s\le e}\Big[\ \min_{e'<s} D_{i-1}(e')\ +\ \phi_i(s,e)\ \Big],
\label{eq:supp-ssa-dp}
\end{equation}
where $D_i(e)$ is the minimum cost of aligning the first $i$ subtitles with
subtitle $i$ ending at window $e$. Back-tracking the minimiser of $D_N$ yields the
aligned spans for all subtitles.

\begin{table}[t]
\centering
\small
\sisetup{round-mode=places, round-precision=2}
\begin{tabular*}{\linewidth}{@{\extracolsep{\fill}}l *{2}{S[table-format=2.2]}}
\toprule
\textbf{Parameter} &
\multicolumn{1}{c}{\textbf{How2Sign}} &
\multicolumn{1}{c}{\textbf{BOBSL}} \\
\midrule
Sliding-window length (frames)            & 12   & 12   \\
Stride (frames)                           & 12   & 4    \\
Candidate range (\texttt{win\_size}, s)   & 26.7 & 26.7 \\
Max inter-sign gap (\texttt{max\_gap}, s) & 8    & 8    \\
Duration weight (\texttt{w\_dur})         & 2    & 2    \\
Inter-sign gap weight (\texttt{w\_gap})   & 2    & 2    \\
Similarity weight (\texttt{w\_sim})       & 5    & 3    \\
Start offset before DP                     & 0.0  & 0.0  \\
End offset before DP                       & 1.0  & 0.0  \\
Start offset after DP                      & 0.0  & 0.0  \\
End offset after DP                        & 0.0  & 1.0  \\
\bottomrule
\end{tabular*}
\caption{\textbf{Best alignment settings per dataset} (offsets and gaps in seconds).}
\label{tab:ssa-params}
\end{table}

\noindent\textbf{Implementation details.}
We embed sliding windows of $12$ frames and aggregate prototype and gloss
similarities by the maximum, sharpening each similarity row by z-scoring before
decoding. Pseudo-glossing uses spaCy~\cite{Honnibal_spaCy} for lemmatisation and part-of-speech
tagging. The decoding weights, candidate range, maximum inter-sign gap, and the
boundary offsets applied before and after the DP are tuned per dataset and listed
in \cref{tab:ssa-params}. The encoder is frozen throughout, so SSA reuses the
exact representation evaluated for retrieval and adds no learnable parameters.
\begin{figure}[t]
    \centering
    \includegraphics[width=1\linewidth]{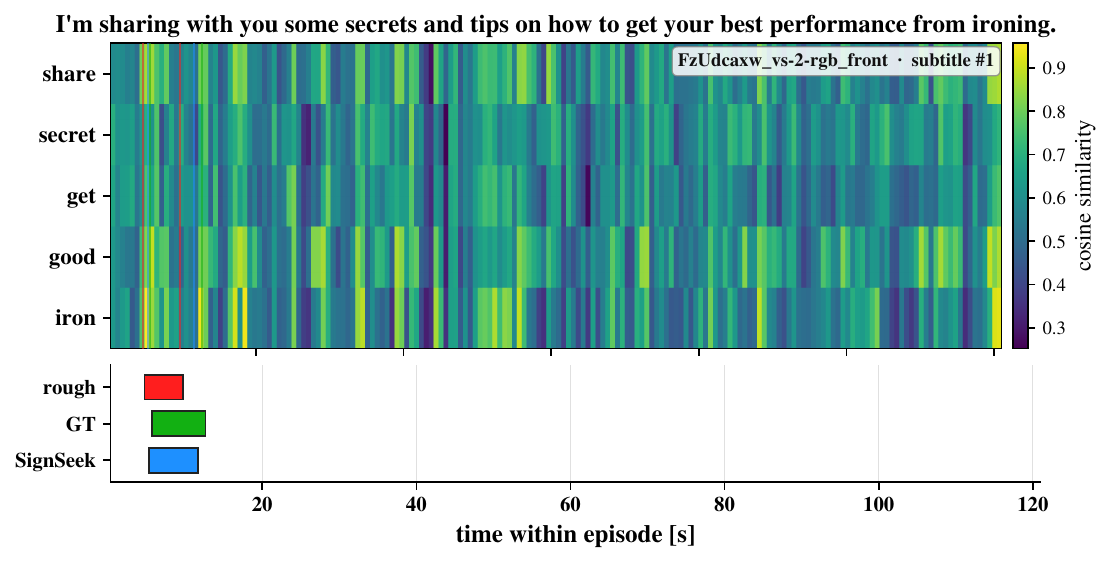}
    \caption{\textbf{Per-subtitle gloss-to-window similarity.} Each row is a
  pseudo-gloss and each column a time window (brighter is more similar). The track
  below shows the rough SRT span, the ground truth, and the \textbf{SignSeek}
  prediction on the same time axis. The decoded span aligns with the
  high-similarity region matching the ground-truth signing.}
    \label{fig:ssa_heatmaps}
\end{figure}
\subsection{Qualitative Results}
\label{sec:supp-ssa-qual}
\begin{figure*}[h]
  \centering
  \begin{subfigure}{\textwidth}
    \includegraphics[width=\linewidth]{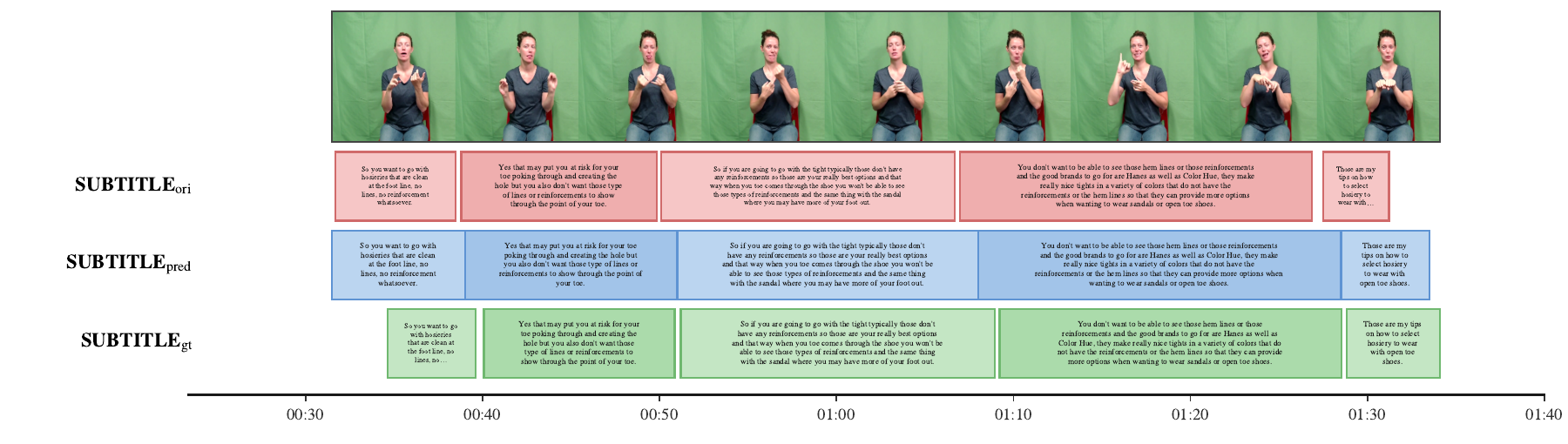}
    \caption{Video ID: \texttt{g1uA0f9I0Sg}.}
    \label{fig:ssa_g3V0BsmDUgY}
  \end{subfigure}\\[6pt]
  \begin{subfigure}{\textwidth}
    \includegraphics[width=\linewidth]{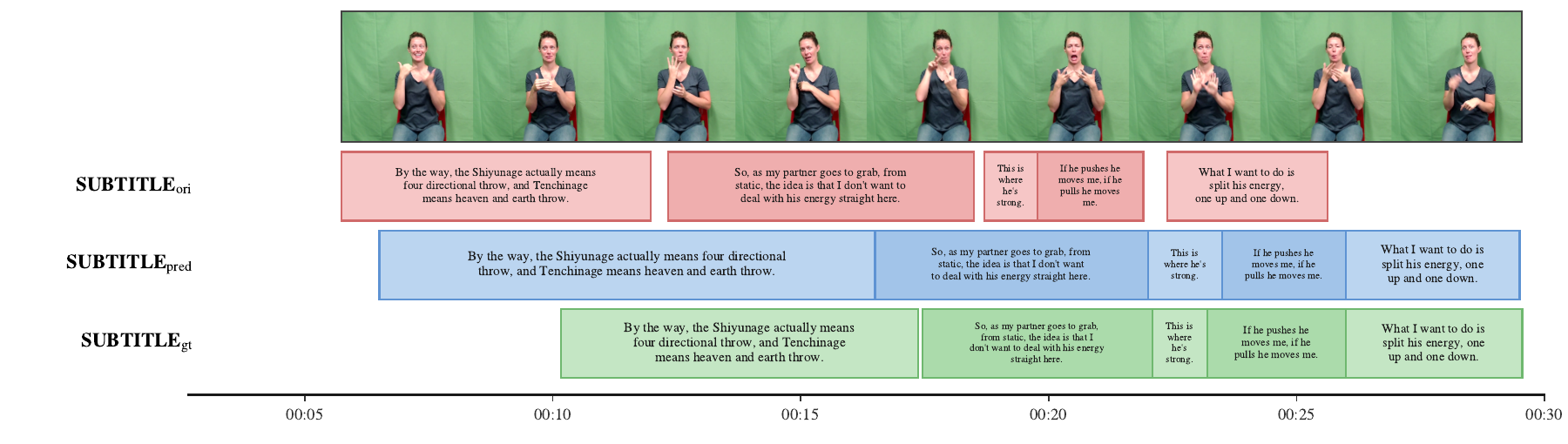}
    \caption{Video ID: \texttt{g3X3XE6M2-A}.}
    \label{fig:ssa_g2SdWBPoXZ0}
  \end{subfigure}\\[6pt]
  \begin{subfigure}{\textwidth}
    \includegraphics[width=\linewidth]{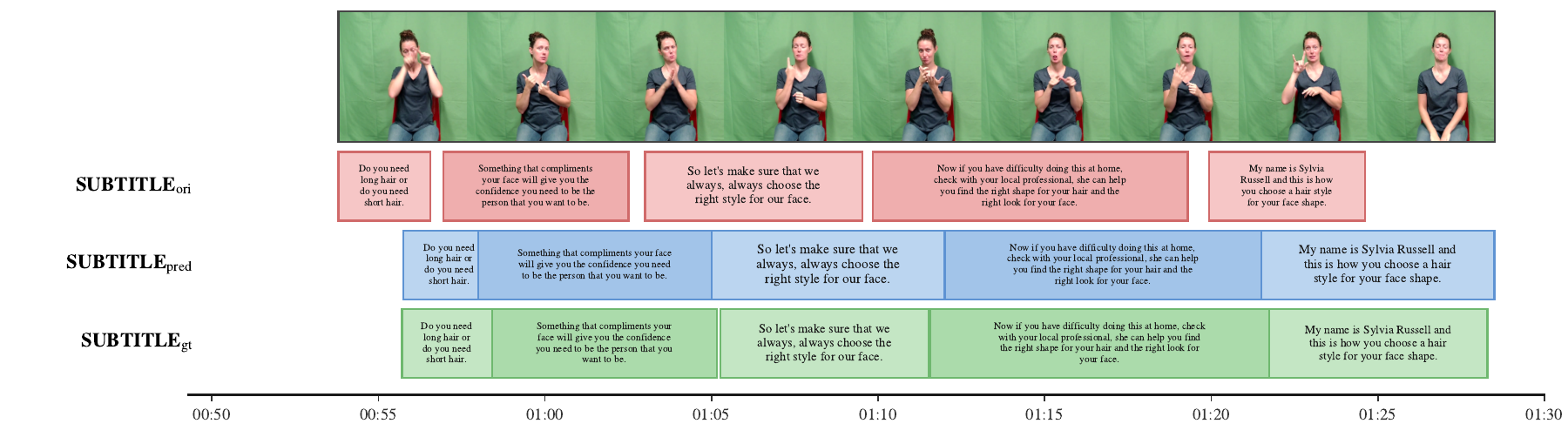}
    \caption{Video ID: \texttt{g3V0BsmDUgY}.}
    \label{fig:ssa_g3ushtMfLiY}
  \end{subfigure}
  \caption{\textbf{Qualitative sign-subtitle alignment on How2Sign.} Original SRT
  (\textrm{ori}), \textbf{SignSeek} prediction (\textrm{pred}), and ground truth
  (\textrm{gt}) spans on a shared timeline, with the RGB filmstrip on top.}
  \label{fig:ssa_qual1}
\end{figure*}

\begin{figure*}[h]
  \centering
  \begin{subfigure}{\textwidth}
    \includegraphics[width=\linewidth]{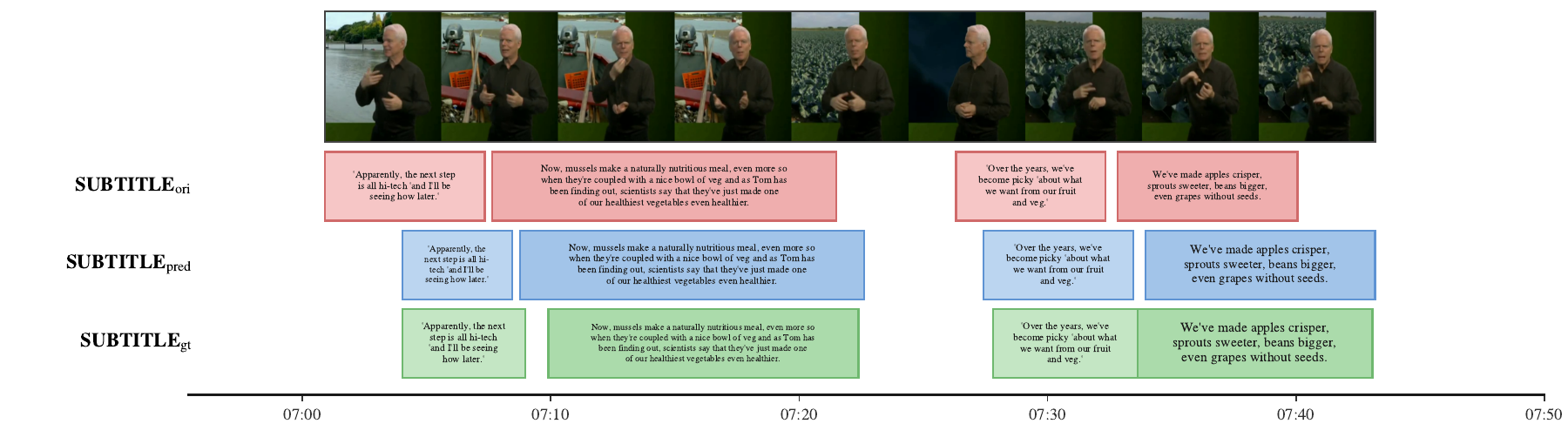}
    \caption{Video ID: \texttt{6212378136168874626}.}
    \label{fig:bobsl_a}
  \end{subfigure}\\[6pt]
  \begin{subfigure}{\textwidth}
    \includegraphics[width=\linewidth]{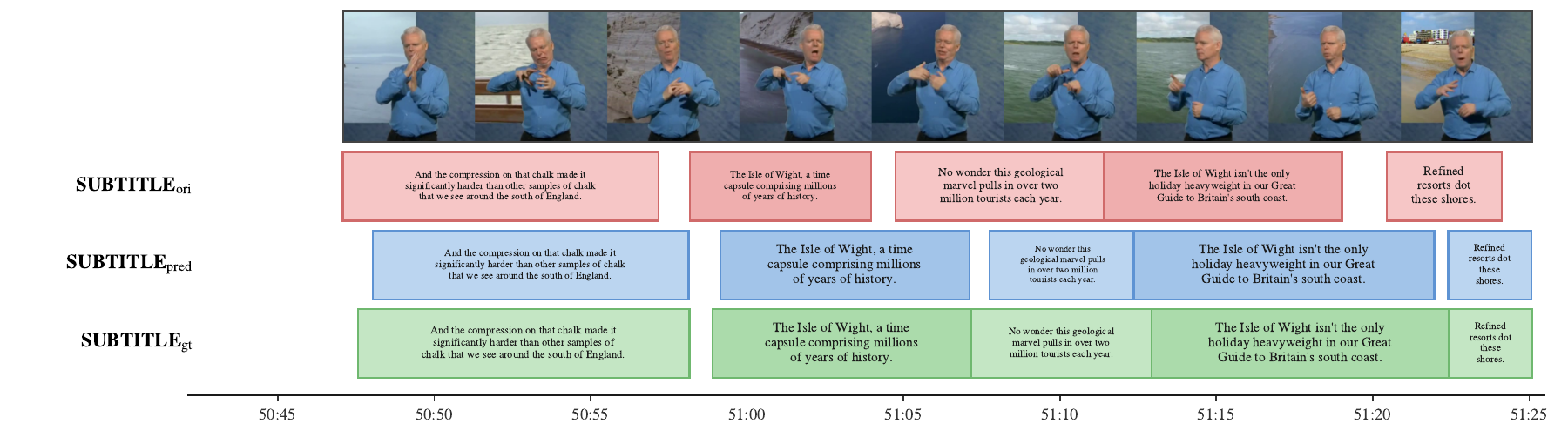}
    \caption{Video ID: \texttt{6358668158244469679}.}
    \label{fig:bobsl_b}
  \end{subfigure}\\[6pt]
  \begin{subfigure}{\textwidth}
    \includegraphics[width=\linewidth]{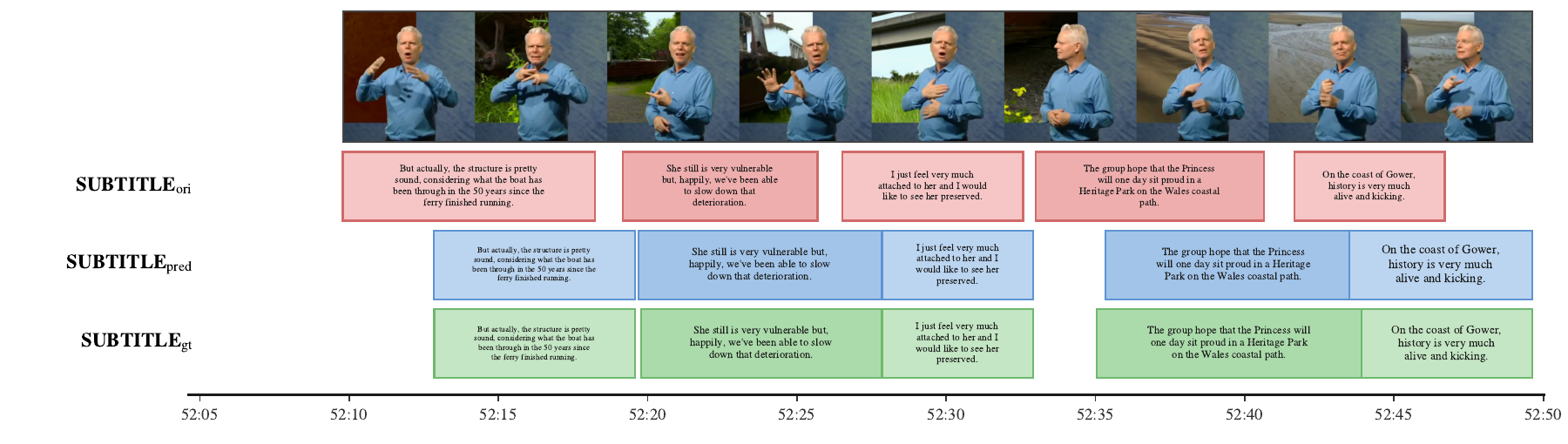}
    \caption{Video ID: \texttt{6342696033864091923}.}
    \label{fig:bobsl_c}
  \end{subfigure}
  \caption{\textbf{Qualitative sign-subtitle alignment on BOBSL (zero-shot BSL).}
  For three episodes we show the original SRT (\textrm{ori}), \textbf{SignSeek}
  prediction (\textrm{pred}), and ground-truth (\textrm{gt}) subtitle spans on a
  shared timeline, with the signer filmstrip on top. Despite never training on BSL,
  \textbf{SignSeek} shifts each subtitle from the drifted SRT timing onto the
  ground-truth signing.}
  \label{fig:ssa_bobsl}
\end{figure*}

\cref{fig:ssa_heatmaps} makes the alignment mechanism explicit for a single
subtitle. Each row is one pseudo-gloss extracted from the subtitle text, and each
column is a sliding window over the episode, with brightness giving the cosine
similarity between the gloss prototypes and that window. The gloss rows light up
together over a short stretch of the timeline, marking where the corresponding
signs are actually produced. The track
below places the rough SRT span, the ground truth, and the \textbf{SignSeek}
prediction on the same axis. The original SRT span sits slightly off the true
signing, and the monotonic decoding moves the subtitle onto the bright
high-similarity region, recovering a span that closely matches the ground truth
without any alignment-specific training.
Beyond this single-subtitle view, \cref{fig:ssa_qual1,fig:ssa_bobsl} show full
episodes on a shared timeline that overlays three tracks, the original SRT spans
(\textrm{ori}), the \textbf{SignSeek} prediction (\textrm{pred}), and the
ground-truth spans (\textrm{gt}), with the signer RGB filmstrip along the top so the
aligned content can be read directly against the video.
\cref{fig:ssa_qual1} shows three How2Sign episodes. The original SRT spans are
offset from the true signing, and \textbf{SignSeek} shifts each subtitle onto the
ground-truth signing, recovering both the onset and the duration of each span
without any alignment-specific training.
\cref{fig:ssa_bobsl} shows three BOBSL episodes. \textbf{SignSeek} shifts the SRT
spans onto the ground-truth signing across the episode.


\clearpage

\end{document}